\pdfoutput=1

\documentclass[11pt]{article}

\usepackage{emnlp2021}

\usepackage{graphicx}
\usepackage{booktabs,tabularx,enumitem,ragged2e}
\usepackage{times}
\usepackage{latexsym}
\usepackage{multirow}
\usepackage{xcolor}
\usepackage{subfigure}
\usepackage{amsmath,amsthm,amssymb}
\usepackage[boldmath]{numprint}

\usepackage{appendix}
\usepackage[T1]{fontenc}

\usepackage[utf8]{inputenc}

\usepackage{microtype}

\newcommand{\data}{\textsc{ExplaGraphs}}

\title{\data: An Explanation Graph Generation Task \\ for Structured Commonsense Reasoning}

\author{Swarnadeep Saha \quad Prateek Yadav \quad Lisa Bauer \quad Mohit Bansal
\\ 
  UNC Chapel Hill\\ 
  \texttt{\{swarna, prateek, lbauer6, mbansal\}@cs.unc.edu}
  \\ 
}

\date{}

\begin{document}
\maketitle
\begin{abstract}
Recent commonsense-reasoning tasks are typically \textit{discriminative} in nature, where a model answers a multiple-choice question for a certain context. Discriminative tasks are limiting because they fail to adequately evaluate the model's ability to reason and explain predictions with underlying commonsense knowledge. They also allow such models to use reasoning shortcuts and not be ``right for the right reasons''. In this work, we present \data{}, a new \textit{generative} and \textit{structured} commonsense-reasoning task (and an associated dataset) of explanation graph generation for stance prediction. Specifically, given a belief and an argument, a model has to predict if the argument supports or counters the belief and also generate a commonsense-augmented graph that serves as non-trivial, complete, and unambiguous explanation for the predicted stance.  We collect explanation graphs through a novel \textit{Create-Verify-And-Refine} graph collection framework that improves the graph quality (up to 90\%) via multiple rounds of verification and refinement. A significant 79\% of our graphs contain external commonsense nodes with diverse structures and reasoning depths. Next, we propose a multi-level evaluation framework, consisting of automatic metrics and human evaluation, that check for the structural and semantic correctness of the generated graphs and their degree of match with ground-truth graphs. Finally, we present several structured, commonsense-augmented, and text generation models as strong starting points for this explanation graph generation task, and observe that there is a large gap with human performance, thereby encouraging future work for this new challenging task.\footnote{\data{} dataset will be publicly available at \url{https://explagraphs.github.io/}.}

\end{abstract}

\begin{figure}[t]
	\centering
    \includegraphics[clip, width=\columnwidth]{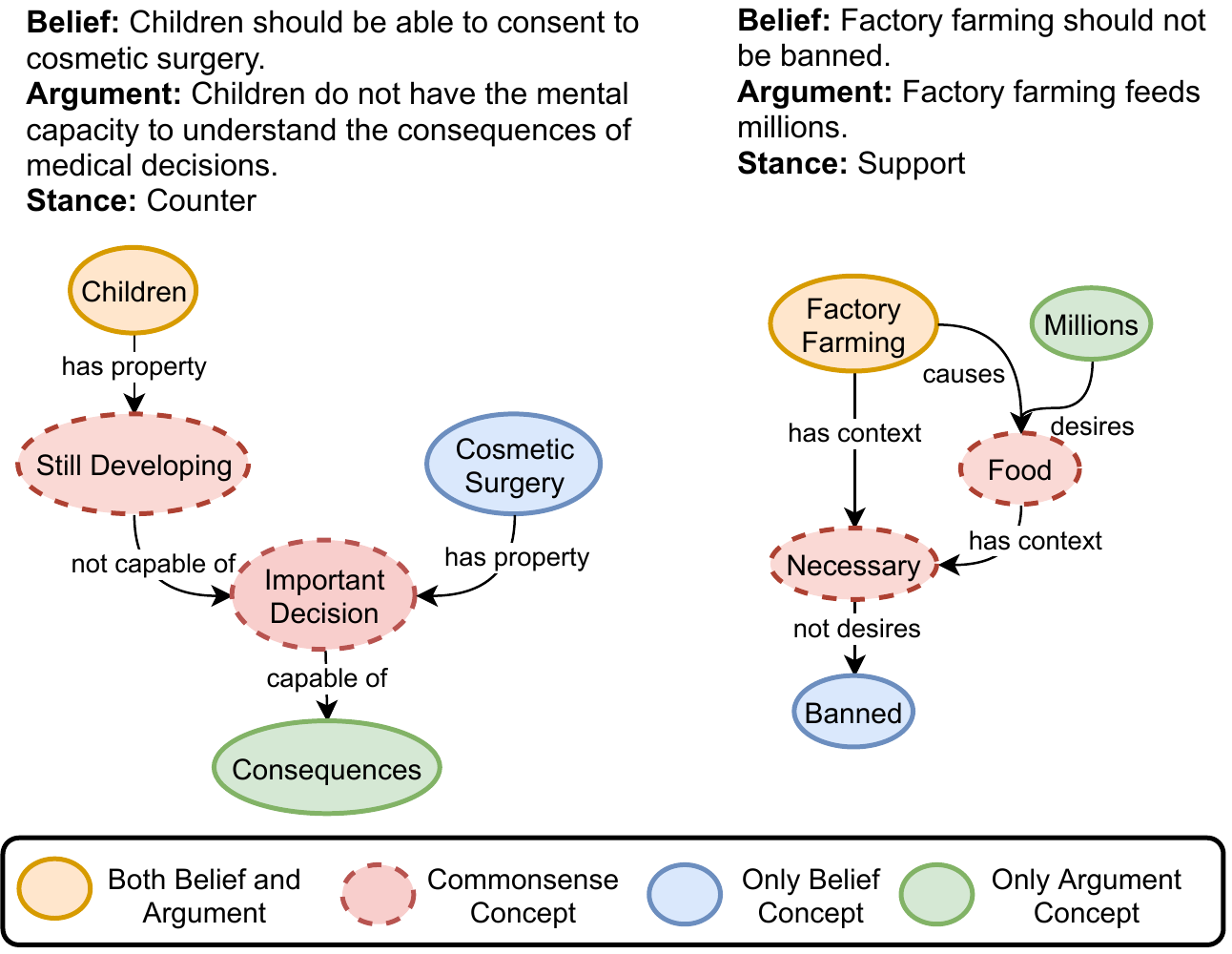}
    \vspace{-15pt}
    \caption{Two representative examples from our dataset. Explanation graphs are read and reasoned through by following the edges that explain why the argument supports or counters the belief. \label{fig:examples_main}
     } 
    \vspace{-5pt}
\end{figure} 

\section{Introduction}
Current state-of-the-art commonsense reasoning (CSR) \cite{davis2015commonsense} models are typically trained and evaluated on \textit{discriminative} tasks, in which a model answers a multiple-choice question for a certain context \cite{zellers2018swag, sap2019socialiqa, bisk2020piqa}. While pre-trained language models perform well on these tasks \cite{lourie2021unicorn}, this setup limits the exploration and evaluation of a model’s ability to reason and explain its predictions with relevant commonsense knowledge, thereby allowing models to solve tasks by using shortcuts, statistical biases or annotation artifacts \cite{gururangan2018annotation, mccoy2019right}. Thus, we emphasize the importance of \textit{generative} CSR capability, in which a model has to compose and reveal the plausible commonsense knowledge required to solve a reasoning task. Moreover, structured (e.g., graph-based) commonsense explanations, unlike unstructured natural language explanations, can more explicitly explain and evaluate the reasoning structures of the model by visually laying out the relevant context and commonsense knowledge edges, chains, and subgraphs. 

We propose \data{}, a new \textit{generative} and \textit{structured} CSR task (in English) of explanation graph generation for stance prediction on debate topics. Specifically, our task requires a model to predict whether a certain argument supports or counters a belief, but correspondingly, also generate a commonsense explanation graph that explicitly lays out the reasoning process involved in inferring the predicted stance. Consider Fig. \ref{fig:examples_main} showing two examples with belief, argument, and stance (support or counter) from our benchmarking dataset collected for this task. Each example requires understanding social, cultural, or taxonomic commonsense knowledge about debate topics in order to infer the correct stance. The example on the left requires the knowledge that ``children'' are ``still developing'' and hence not capable of making an ``important decision'' like ``cosmetic surgery'' which has ``consequences''. Given this knowledge, one can understand that the argument is counter to the belief. We represent this knowledge in the form of a commonsense explanation graph.

Graphs are efficient for representing explanations due to multiple reasons: (1) unlike a chain of facts \cite{khot2020qasc, jhamtani2020learning, inoue2020r4c, geva2021strategyqa}, they can capture complex dependencies between facts, while also avoiding redundancy (e.g., ``Factory farming causes food and millions desire food'' forms a ``V-structure''), (2) unlike natural language explanations \cite{camburu2018snli, rajani2019explain, narang2020wt5, brahman2020learning, zhang2020winowhy}, it is easier to impose task-specific constraints on graphs (e.g., connectivity, acyclicity), that eventually help in better quality control during data collection (Sec. \ref{sec:dataset}) and designing structural validity metrics for model-evaluation (Sec. \ref{sec:metrics}), and (3) unlike semi-structured templates \cite{ye2020teaching, mostafazadeh2020glucose} or extractive rationales \cite{zaidan2007using, lei2016rationalizing, yu2019rethinking, deyoung2020eraser}, they allow for more flexibility and expressiveness. Graphs can encode any reasoning structure and the nodes are not limited to just phrases from the context. As shown in Fig. \ref{fig:examples_main}, our explanations are connected directed acyclic graphs (DAGs), in which the nodes are either internal concepts (short phrases from the belief or argument), or external commonsense concepts (dashed-red), essential for connecting the internal concepts in a way that the stance is inferred. The edges are labeled with commonsense relations chosen from a pre-defined set. While some edges might not necessarily be factual (e.g., ``Factory farming; has context; necessary''), note that such edges are essential in the context for composing an explanation that is indicative of the stance. Semantically, our graphs are extended structured arguments, augmented with commonsense knowledge.

We construct a benchmarking dataset for our task through a novel \textit{Create-Verify-And-Refine} graph collection framework. These graphs serve as non-trivial (not paraphrasing the belief as an edge), complete (explicitly connects the argument to the belief) and unambiguous (infers the target stance) explanations for the task (Sec. \ref{sec:task_definition}). The graph quality is iteratively improved (up to 90\%) through multiple verification and refinement rounds. 79\% of our graphs contain external commonsense nodes, indicating that commonsense is a critical component of our task. Explanation graph generation poses several syntactic and semantic challenges like predicting the internal nodes, generating the external concepts and predicting and labeling the edges in a way that leads to a connected DAG. Finally, the graph should unambiguously infer the target stance.

We next present a multi-level evaluation framework for our task (Sec. \ref{sec:metrics}, Fig. \ref{fig:metrics}), consisting of diverse automatic metrics and human evaluation. The evaluation framework checks for stance and graph consistency along with the structural and semantic correctness of explanation graphs, both locally by evaluating the importance of each edge and globally by the graph's ability to reveal the target stance. Furthermore, we propose graph-matching metrics like Graph Edit Distance \cite{abu2015exact} and ones that extend text-generation metrics for graphs (based on multiple test graphs in our dataset). Lastly, as some strong initial baseline models for this new task, we propose a commonsense-augmented structured prediction model that predicts nodes and edges jointly and enforces global graph constraints (e.g., connectivity) through an Integer Linear Program (ILP). We also experiment with BART \cite{lewis2019bart} and T5 \cite{2020t5} based models, and show that all these models have difficulty in generating meaningful graph explanations for our challenging task, leaving a large gap between model and human performance. Overall, our main contributions are:
\begin{itemize}[nosep, wide=0pt, leftmargin=*, after=\strut]
    \item We propose \data{}, a \textit{generative} and \textit{structured} commonsense-reasoning task of explanation graph generation for stance prediction.
    \item We construct a benchmarking dataset for our task and propose a novel \textit{Create-Verify-And-Refine} graph collection framework for collecting graphs that serve as explanations for the task. Our framework is generalizable to any crowdsourced collection of graph-structured data.
    \item We propose a multi-level evaluation framework with automatic metrics and human evaluation, that compute structural and semantic correctness of graphs and match with human-written graphs. 
    \item We propose a commonsense-augmented structured model and BART/T5 based models for this task, and find that they are relatively weak at generating reasoning graphs, obtaining 20\% accuracy (compared to human performance of 84\%).
\end{itemize}{}
\vspace{-7pt}
We encourage researchers to use our benchmark as a way to improve and explore structured commonsense reasoning capabilities of models.

\section{Related Work}
\noindent \textbf{Structured Explanations in NLP:} Explanation datasets in NLP \cite{wiegreffe2021teach} take three major forms: (1) extractive rationales \cite{zaidan2007using, lei2016rationalizing, yu2019rethinking, deyoung2020eraser}, (2) free-form or natural language explanations \cite{camburu2018snli, rajani2019explain, narang2020wt5, brahman2020learning, zhang2020winowhy}, and (3) structured explanations consisting of explanations graphs \cite{jansen2018worldtree, jansen2019textgraphs, xie2020worldtree, saha2020prover, kalyanpur2020braid, saha2021multiprover}, chain of facts \cite{khot2020qasc, jhamtani2020learning, inoue2020r4c, geva2021strategyqa} or semi-structured text \cite{ye2020teaching}. Our commonsense explanations bear most similarity to WorldTree's \cite{jansen2018worldtree} explanation graphs. However, while they connect lexically-overlapping words, we connect concepts to create facts and diverse reasoning structures in fully-structured graphs with carefully designed constraints for explainability. \data{}'s explanations also share similarities with visual scene graphs from the vision community \cite{johnson2015image, xu2017scene}, in which the image entities are connected via edges to represent relationships.

\begin{figure*}[t]
	\centering
    \includegraphics[clip, width=\textwidth]{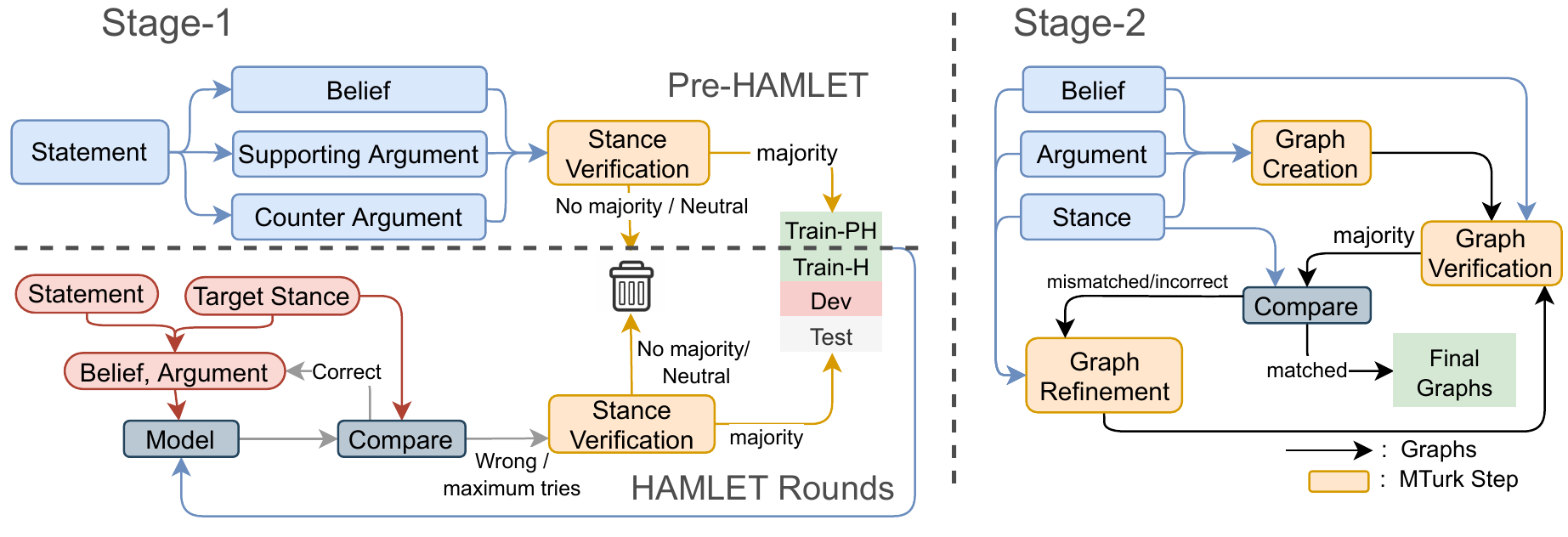}
    \vspace{-20pt}
     \caption{Interface for our data collection framework consisting of two stages. In Stage 1, we collect (belief, argument, stance) triples in pre-HAMLET and multiple HAMLET (human-and-model-in-the-loop) rounds. In each HAMLET round, we collect harder examples by asking the annotators to fool a stance prediction model. In Stage 2, we collect the corresponding explanation graphs through a Create-Verify-And-Refine framework.\label{fig:steps}} 
     \vspace{-3pt}
\end{figure*}

\noindent \textbf{Commonsense Reasoning Benchmarks:} A large variety of CSR tasks have been developed recently, including commonsense extraction \cite{li2016commonsense, xu2018automatic}, next situation prediction \cite{zellers2018swag, zellers2019hellaswag}, cultural, social, and physical commonsense understanding \cite{lin2018mining, sap2019atomic, sap2019socialiqa, bisk2020piqa, hwang2020comet, forbes2020social}, pronoun disambiguation \cite{sakaguchi2020winogrande, zhang2020winowhy}, abductive commonsense reasoning \cite{bhagavatula2019abductive} and general commonsense \cite{talmor2019commonsenseqa, huang2019cosmos, wang2019does, boratko2020protoqa}. While there is an abundance of discriminative commonsense tasks, there are few recent works in generative commonsense tasks. E.g., CommonGen \cite{lin2020commongen} generates unstructured commonsense text, and EIGEN \cite{madaan2020eigen} considers event influence graph generation. Instead, our work focuses on generating commonsense-augmented explanation graphs.

\noindent \textbf{Stance Prediction and Argumentation:} Previous stance prediction works have been largely applied to online content, for political, ideological debates, rumor and fake news detection \cite{mohammad2016semeval, derczynski2017semeval, hardalov2021survey}. Other recent works on argumentation deal with convincingness of claims and arguments \newcite{habernal2016argument, gleize2019you} and reasons \cite{hasan2014you}. However, to the best of our knowledge, our work is the first to explore explicit commonsense-augmented graph-based explanations for stance prediction.

\section{\data{} Task Definition}
\label{sec:task_definition}
We propose \data{}, a new \textit{generative} and \textit{structured} commonsense-reasoning task, where given a belief about a topic and an argument, a model has to (1) infer the stance (support/counter), and (2) generate the corresponding commonsense explanation graph that explains the inferred stance (Fig. \ref{fig:examples_main}). Our primary focus in this work is on the second sub-task that requires generative commonsense reasoning. The explanation graph is a connected and directed acyclic graph, where each node is a concept (short English phrase). Concepts are either internal (part of the belief or the argument) or external (part of neither but essential for filling in any knowledge gap between the belief and the argument). Each directed edge connects two concepts and is labeled with one of the pre-defined commonsense relations. These relations are chosen based on ConceptNet \cite{liu2004conceptnet} with three modifications -- (1) removing some generic relations like “related to”, (2) merging some relations that have similar meanings (e.g. ``synonym of'' and ``similar to''), (3) adding a negated counterpart (``not desires'') for every non-negated relation (``desires''), to enable easy construction of support and counter explanations and a balanced set between negated and non-negated relations (see appendix for full list). Semantically, our explanation graphs are commonsense-augmented structured arguments that explicitly support or counter the belief. All subjective claims in the graph are assumed to be true for inferring the stance. An explanation graph is correct if it is both structurally and semantically correct.

\paragraph{Structural Correctness of Graphs: }In order to ensure the structural validity of an explanation graph, we define certain constraints on the graph which not only ensure better quality control during our data collection (Sec. \ref{sec:dataset}) but also simplify the evaluation (Sec. \ref{sec:metrics}), given the open-ended nature of our task. Note that most of these constraints are only possible to impose because of the explicit graphical structure of these explanations.

\begin{itemize}[nosep, wide=0pt, leftmargin=*, after=\strut]
    \item Each concept should contain a maximum of three words and each relation should be chosen from the pre-defined set of relations.
    \item The total number of edges should be between 3 and 8, to ensure a good balance between under-specified and over-specified explanations.
    \item The graph should contain at least two concepts from the belief and at least two from the argument. This ensures that the graph uses important parts of the belief and argument (exactly, without paraphrasing) to construct the explanation.
    \item The graph should be a connected DAG to ensure the presence of explicit reasoning chains between the belief and argument and also avoid redundancy or circular explanations. E.g., having ``(vegans; antonym of; meat eaters)'' makes ``(meat eaters; antonym of; vegans)'' redundant.
\end{itemize}{}

\paragraph{Semantic Correctness of Graphs: }We define the semantic correctness of explanation graphs as follows. First, all facts in the graph, individually, should be \textit{semantically coherent}. Second, the graph should be \textit{non-trivial}, \textit{complete} and \textit{unambiguous}. We call a graph \textit{non-trivial} if it uses the argument to arrive at the belief and does not use fact(s) which are mere paraphrases of the belief. E.g., for a belief ``Factory farming should be banned'', if the explanation graph contains facts like ``(Factory farming; desires; banned)'', then it is only paraphrasing the belief to explain why the belief holds, hence making the graph incorrect. Instead, it should be augmenting the argument with commonsense knowledge like our graph in Fig. \ref{fig:examples_main}. A \textit{complete} graph is one which explicitly connects the argument to the belief and no other commonsense knowledge is needed to understand why it supports or counters the belief. E.g., in Fig. \ref{fig:examples_main}, the fact ``(necessary; not desires; banned)'' makes the explanation complete by explicitly connecting back to the belief. We call a graph \textit{unambiguous} if it, as a whole, infers the target stance and only that stance. We revisit these definitions of structural and semantic correctness when evaluating the quality of human-written graphs (Sec. \ref{sec:graph_collection}) as well as model-generated graphs (Sec. \ref{sec:metrics}).

\section{Dataset Collection}
\label{sec:dataset}
We collect \data{} data in two stages via crowdsourcing on Amazon Mechanical Turk (Fig. \ref{fig:steps}). In Stage 1 (left of Fig. \ref{fig:steps}), we collect instances of belief, argument and their corresponding stance. In Stage 2 (right of Fig. \ref{fig:steps}), we collect the corresponding commonsense explanation graph for each (belief, argument, stance) sample.

\subsection{Stage 1: (Belief, Argument, Stance)}
In Stage 1, annotators are given prompts that express beliefs about various debate topics, extracted from evidences in \citet{gretz2019large}. We use 71 topics in total (see appendix for the list), randomly assigning 53/9/9 disjoint topics to our train/dev/test splits. Given the prompt, annotators write the belief expressed in the prompt and subsequently, a supporting and a counter argument for the belief. Since we focus on commonsense-augmented explanations, we want to ensure that most of our belief, argument pairs require some implicit background commonsense knowledge for understanding why a certain argument supports or refutes the belief. For collecting such pairs, we use Human-And-Model-in-the-Loop Enabled Training (HAMLET) \cite{nie2019adversarial}, a multi-round adversarial data collection procedure that enables the collection of trickier examples with more background commonsense knowledge. Due to space constraints, we discuss this in detail in appendix Sec. 1.1. After stance label verification, we obtain a high fleiss-kappa inter-annotator agreement of $0.61$.

\subsection{Stage 2: Commonsense Explanation Graph Collection}
\label{sec:graph_collection}
Given the (belief, argument, stance) triples from Stage 1, we next collect the corresponding commonsense explanation graphs through a generic \textit{Create-Verify-And-Refine} iterative framework. 

\begin{figure}[t]
	\centering
    \includegraphics[width=\columnwidth]{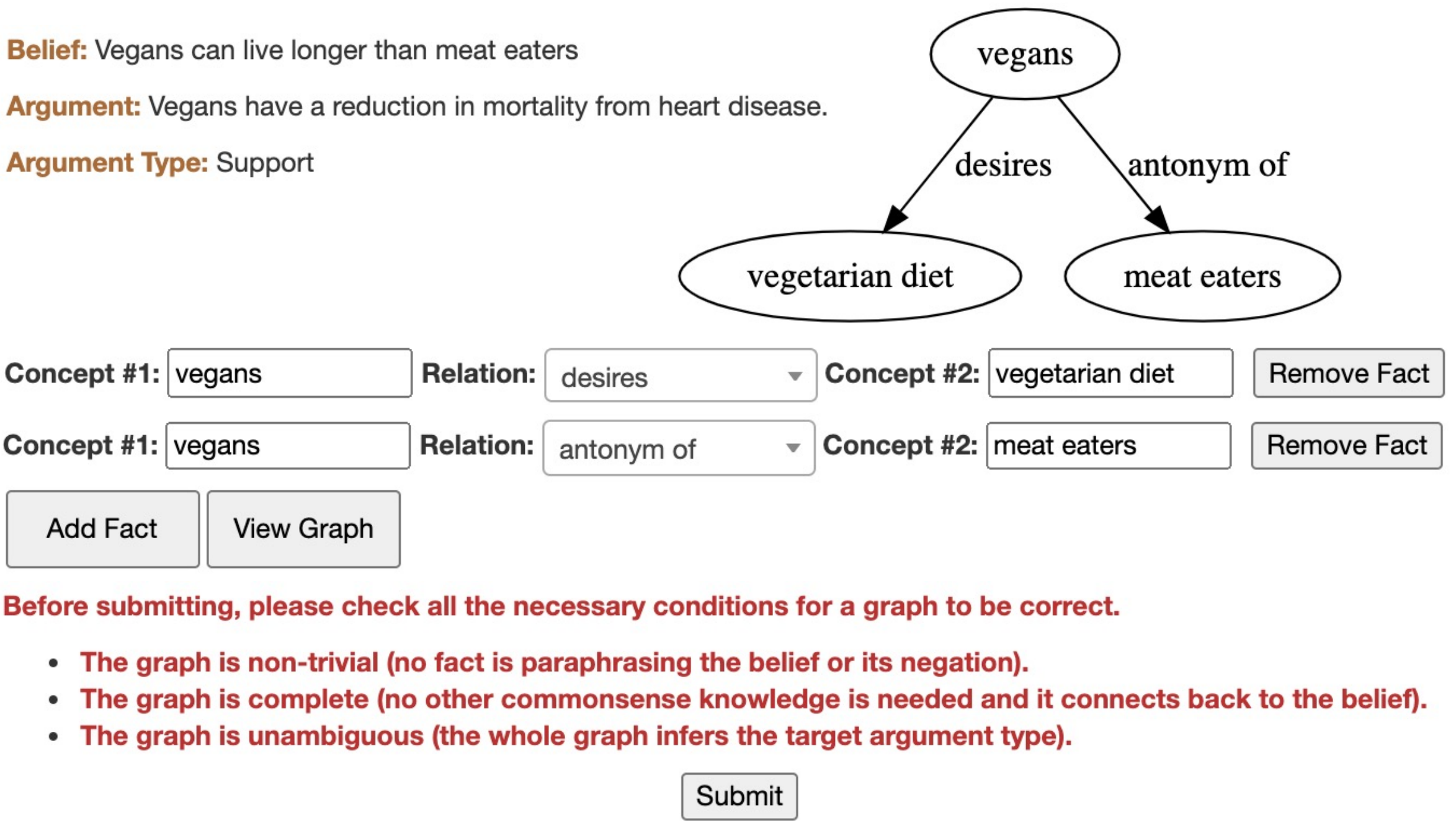}
    \vspace{-15pt}
     \caption{Explanation Graph Creation Interface.
     }
     \vspace{-10pt}
     \label{fig:stage2_interface}
\end{figure}

\noindent \textbf{Graph Creation:}
Annotators are given a belief, an argument, and the stance (support or counter) and are asked to construct a commonsense-augmented explanation graph that explicitly explains the stance (Fig. \ref{fig:stage2_interface}). A graph is constructed by writing multiple facts, each consisting of two concepts and a chosen relation that connects the two concepts. The annotators write 3-8 facts such that the facts lead to a connected DAG with at least two concepts from the belief and two from the argument. The graphical representation of the explanation provides an explicit structure, thereby allowing us to automatically perform in-browser checks for these structural constraints. Clicking on the ``View Graph'' button shows the graph written so far. Before submitting, we remind the annotators that they reason through the graph and verify that it is non-trivial, complete and unambiguous (marked red in Fig. \ref{fig:stage2_interface}). See appendix for the graph creation instructions.

\noindent \textbf{Graph Verification:}
Here, we verify the semantic correctness of graphs (as defined in Sec. \ref{sec:task_definition}) because by construction, they are all structurally correct. The explanation graphs should be complete and hence are treated as extended structured arguments with commonsense. Thus, in our graph verification step, we provide annotators with only the belief and the corresponding explanation graph and ask them to reason through it to infer the stance. Additionally, we include a third category of ``incorrect'' graphs which is broadly aimed at identifying the ill-formed graphs with either semantically incoherent facts, trivial belief-paraphrased facts, or no explicit connection back to the belief (incomplete or ambiguous). Each graph is annotated by three verifiers into one of support/counter/incorrect. A graph is considered correct if and only if the majority label matches the original stance (already known from Stage 1). All other graphs are sent for refinement (described next, also see Fig. \ref{fig:steps}) because they are either incorrect or infer the wrong stance. See appendix for the graph verification interface.

\begin{table*}[tbh]
\small
\centering
\resizebox{1.0\textwidth}{!}{\begin{minipage}{\textwidth}
\centering
\begin{tabular}{lccccccccc}
\toprule
           & \multicolumn{3}{c}{\textbf{Train}}           & \multicolumn{3}{c}{\textbf{Dev}}         & \multicolumn{3}{c}{\textbf{Test (2 graphs/sample)}}         \\ 
           \textbf{Round} & S / C          & Total & Topics & S / C      & Total & Topics & S / C       & Total & Topics \\ \midrule
Pre-HAMLET & 541 / 457  & 998  & 33     & -              & -     &   -     & -               & -     &      -  \\
HAMLET R1  & 347 / 226    & 573   & 20     & 79 / 76   & 155   & 9      & 84 / 80    & 164   & 9      \\
HAMLET R2  & 234 / 181    & 415   & 20     & 66 / 63   & 129   & 9      & 64 / 59    & 123   & 9      \\
HAMLET R3  & 213 / 169    & 382   & 20     & 54 / 60   & 114   & 9      & 52 / 61    & 113   & 9      \\ \midrule
\data{} & 1335 / 1033 & 2368  & 53     & 199 / 199 & 398   & 9      & 200 / 200 & 400   & 9 \\ \bottomrule    
\end{tabular}
\end{minipage}}
\vspace{-5pt}
\caption{\data{} dataset statistics: S = Support, C = Counter, Topics = Number of spanning topics.}
\vspace{-5pt}
\label{tab:dataset}

\end{table*}

\begin{table}[t]
\small
    \centering
    \resizebox{0.99\columnwidth}{!}{\begin{minipage}{\columnwidth}
    \begin{tabular}{ccccccc}
    \toprule
         & \#N & \#E & \#EN & D & \%Non-linear & \%EN \\ \midrule
         Train & 5.1 & 4.2 & 1.3 & 3.3 & 58.8 & 78.2 \\ 
         Dev & 5.4 & 4.5 & 1.6 & 3.8 & 47.0 & 88.4 \\
         Test & 5.2 & 4.3 & 1.4 & 3.3 & 63.9 & 78.4 \\ \midrule
         Total & 5.2 & 4.3 & 1.3 & 3.4 & 58.6 & 79.4 \\ \bottomrule
    \end{tabular}
    \end{minipage}}
    \vspace{-2pt}
    \caption{Graph Statistics: \#N, \#E, \#EN = Average number of nodes, edges and external nodes respectively. D = Average depth of graphs. \%Non-Linear = percentage of graphs which are not linear chains. \% EN = percentage of graphs with external node(s) in the graph.}
    \vspace{-5pt}
    \label{tab:graph_stats}
\end{table}

\noindent \textbf{Graph Refinement:}
During graph refinement, in addition to the belief, argument, and the target stance, annotators are provided with the initial incorrect graph along with the verification label from the previous stage. Then another qualified annotator who is not the author of the initial graph is asked to refine it. Refinement is defined in terms of three edit operations on the graph: (1) adding a new fact, (2) removing an existing fact, and (3) replacing an existing fact. We again ensure that the refined graph adheres to the structural constraints. See appendix for the instructions and interface.

\noindent \textbf{Graph Quality:} The refined graphs are again sent to the verification stage and the process iterates between the verification and refinement stages until we obtain a high percentage of correct graphs. We perform two rounds of refinement, and obtain a high \emph{90\% of semantically correct graphs} (67\%, 81\% and 90\% after rounds 1, 2 and 3 respectively). Our \textit{Create-Verify-And-Refine} framework is generic and allows for iterative improvement of graphs. See appendix for various quality control mechanisms for complex graph collection, which we believe will be helpful for similar future efforts.
    
\section{Dataset Analysis}
\label{sec:dataset-details}

\data{} consists of a total of 3166 samples (see Table \ref{tab:dataset}).\footnote{Like prior structured data collection efforts \cite{geva2021strategyqa}, graph collection is challenging due to the difficulty in training annotators to create (connected/acyclic) graphs and verifying them for semantic consistency and stance inference.} 
We collect two graphs for each sample in the test set. Table \ref{tab:graph_stats} shows statistics concerning the average number of nodes, edges, and external commonsense nodes present in our graphs. Approximately, 79\% of graphs contain external nodes, indicating that most of our samples require background commonsense knowledge to explicitly support or refute a belief. Additionally, our graphs have diverse reasoning structures, with 58\% of non-linear graphs. A large presence of non-linear structures and an average depth of 4 indicates complex reasoning involved in our task. We also find that the most frequently used relations are causal (like ``capable of'', ``causes'', ``desires'', and their negative counterparts), which further supports our graphs as explanations (details in appendix).

\section{Evaluation Metrics}
\label{sec:metrics}
\begin{figure}
\centering
    \includegraphics[clip, width=0.95\columnwidth]{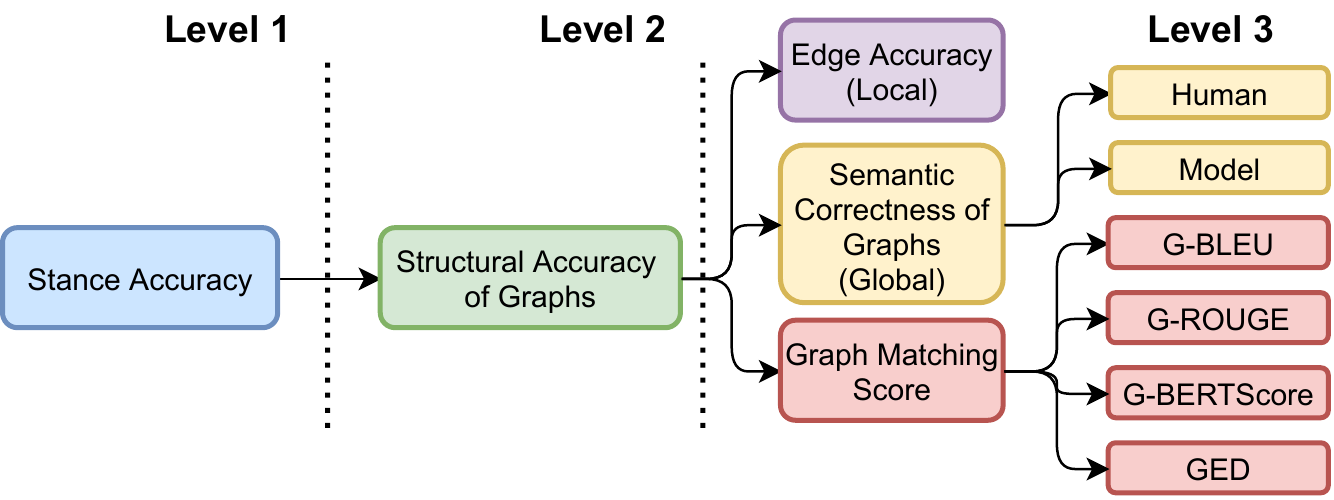}
    \vspace{-5pt}
    \caption{Our multi-level evaluation framework.
    \vspace{-10pt}
    \label{fig:metrics}}
\end{figure}

Explanation graphs can be represented in multiple correct ways with varying levels of specificity and different graphical structures. A single concept can also be paraphrased differently. Thus, we design a 3-level evaluation pipeline (see Fig. \ref{fig:metrics}).

\vspace{-2pt}
\paragraph{Level 1 -- Stance Accuracy (SA):} All models for our task predict both the stance label and the commonsense explanation graph. In Level 1, we report the stance prediction accuracy which ensures that the explanation graph is consistent with the predicted stance. Samples with a correctly predicted stance are then passed to the next levels that check for the quality of the generated explanation graphs.

\vspace{-2pt}
\paragraph{Level 2 -- Structural Correctness Accuracy of Graphs (StCA):} As per our task definition in Sec. \ref{sec:task_definition}, for an explanation graph to be correct, it first has to be structurally correct. Hence, we compute the fraction of structurally correct graphs (connected DAGs with at least three edges and at least two concepts from the belief and at least two from the argument). Samples with correct stances and structurally correct graphs are then evaluated in Level 3 for: (1) semantic correctness, (2) match with GT graphs, and (3) edge importance. 

\vspace{-2pt}
\paragraph{Level 3 -- Semantic Correctness Accuracy of Graphs (SeCA):} Identifying semantic correctness of a graph requires following our human verification process discussed in Sec. 4.2. A graph is semantically correct if all its edges are semantically coherent and given the belief, the unambiguously inferred stance from the graph matches the original stance. However, both these aspects are challenging because they require understanding the underlying semantics and reasoning through the graph. Carrying this out by humans at a large scale is also expensive. Thus, following previous works \cite{bertscore,sellam2020bleurt,pruthi2020evaluating}, we propose an automatic model-based metric that given a belief-graph pair, predicts the label between incorrect, support, and counter. Specifically, we fine-tune RoBERTa \cite{liu2019roberta} on the beliefs and corresponding human-verified graphs from our data collection phase. Graphs are fed as concatenated edges to the model. Since the space of incorrect graphs is potentially huge, we augment our training data with synthetically created incorrect graphs (by randomly adding, removing, or replacing edges) from already correct (support/counter) graphs. Note that our automatic metric is not meant to replace human evaluation. Thus, for completeness, we still perform human evaluation and show human-metric correlation for SeCA (Sec. \ref{sec:expt}).

\vspace{-2pt}
\paragraph{Level 3 -- G-BERTScore (G-BS):} We also introduce a matching metric that quantifies the degree of match between the ground-truth and the predicted graphs. We call this G-BERTScore, designed as an extension of a text generation metric, BERTScore \cite{bertscore} for graph-matching. We consider graphs as a set of edges and solve a matching problem that finds the best assignment between the edges in the gold graph and those in the predicted graph. Each edge is treated as a sentence and the scoring function between a pair of gold and predicted edges is given by BERTScore.\footnote{We choose BERTScore over BLEU or ROUGE because they have been shown to correlate poorly with humans for prior natural language explanation studies \cite{camburu2018snli, marasovic2020natural}. However, for completeness sake, our code reports them.} Given the best assignment and the overall matching score, we compute precision, recall and report F1 as our G-BERTScore metric. On the test set, we consider the best match across all ground-truth graphs.

\paragraph{Level 3 -- Graph Edit Distance (GED): } As a more interpretable graph matching metric, we use Graph Edit Distance \cite{abu2015exact} to compute the distance between the predicted graph and the gold graph. Formally, GED measures the number of edit operations (addition, deletion, and replacement of nodes and edges) for transforming the predicted graph to a graph isomorphic to the gold graph. The cost of each edit operation is chosen to be 1. The GED for each sample is normalized between 0 and 1 by an appropriate normalizing constant (upper bound of GED). Thus, the samples with either incorrect stances or structurally incorrect graphs will have a maximum normalized GED of 1 while samples whose graphs match exactly will have a score of 0. The overall GED is given by the average of the sample-wise GEDs. Lower GED indicates that the predicted graphs match more closely with the gold graphs.

\begin{figure}
\centering
    \includegraphics[clip, width=0.99\columnwidth]{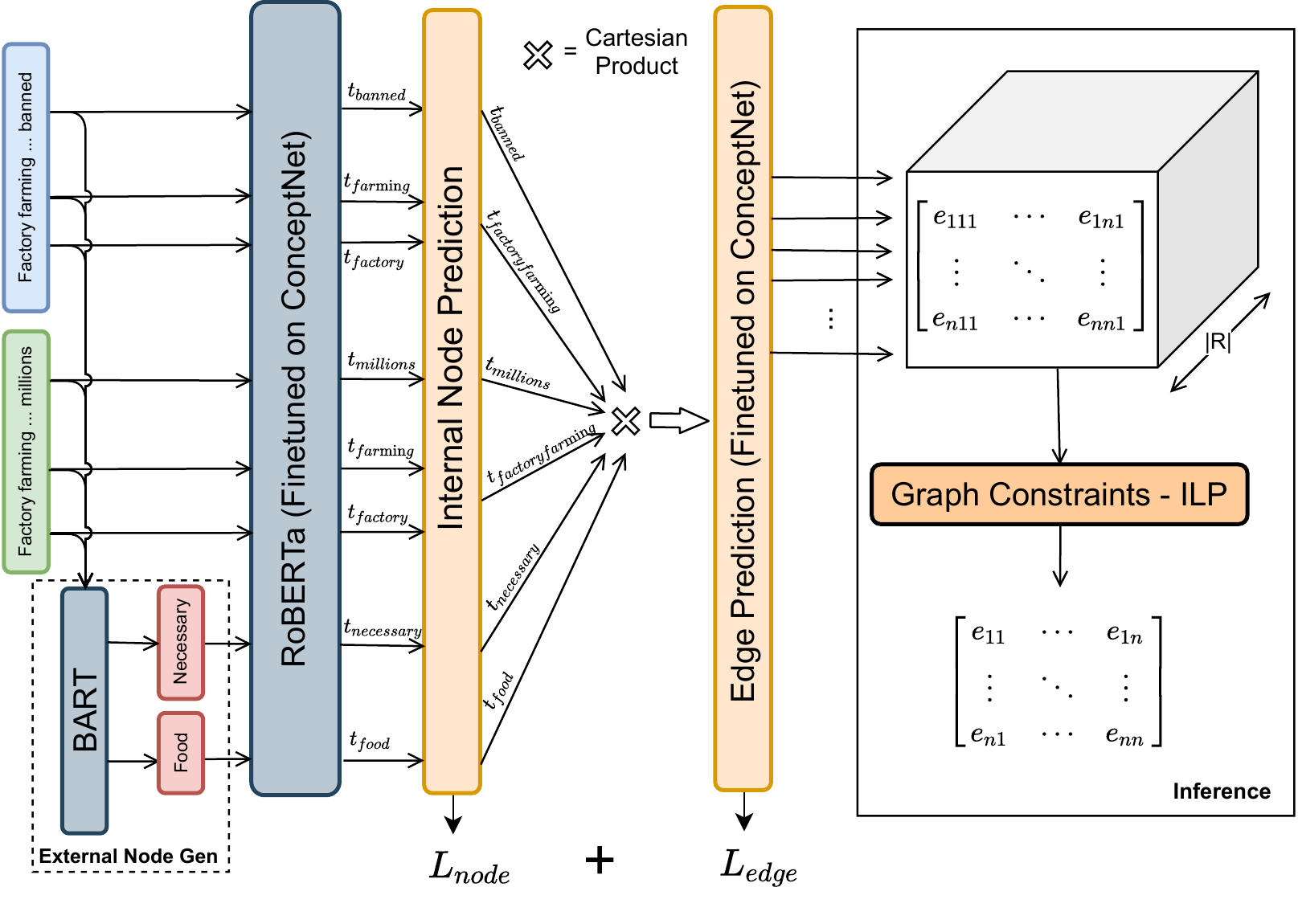}
    \vspace{-17pt}
    \caption{Our Commonsense-Augmented Structured Prediction Model for explanation graph generation. \label{fig:model}}
    \vspace{-7pt}
\end{figure} 

\begin{table*}[t]
\small
\centering
\resizebox{0.85\textwidth}{!}{
\begin{tabular}{lccccccc}
\toprule
& SA$\uparrow$ \;\;\; & StCA$\uparrow$ \;\;\; & SeCA$\uparrow$ \;\;\; & G-BS$\uparrow$ \;\;\; & GED$\downarrow$ \;\;\; & EA$\uparrow$ \;\;\;\;\; & Hu$\uparrow$ \\ \midrule
RE-BA \;\;\;\;\;\; & 68.5 & 18.7 & 11.2 & 15.6 & 0.86 & 10.0 & 5.7 \\
RA-BA & \textbf{87.2} & 25.7 & 13.0 & 22.0 & 0.81 & 12.8 & 8.5 \\
RE-T5 & 69.0 & 32.5 & 13.5 & 28.3 & 0.75 & 17.3 & 8.7 \\
RA-T5 & \textbf{87.2} & 38.7 & 19.0 & 33.6 & 0.71 & 20.8 & 10.5  \\
RE-SP & 72.5 & \textbf{62.5} & \textbf{20.0} & \textbf{50.0} & \textbf{0.60} & \textbf{26.2} & \textbf{12.5}\\
\midrule
UB & 91.0 & 91.0 & 83.5 & 71.1 & 0.38 & 46.8 & 80.3
\\\bottomrule                                             
\end{tabular}}
\vspace{-3pt}
\caption{Results of our models across all metrics on \data{} test set. UB = Metric Upper Bound, Hu = Human verification of semantic correctness of graphs.
}
\label{tab:model_test}

\vspace{-7pt}
\end{table*}

\vspace{-2pt}
\paragraph{Level 3 -- Edge Importance Accuracy (EA):}
While SeCA assesses the correctness of a graph at a global level, we also propose a local model-based metric, named ``Edge Importance Accuracy'' which computes the macro-average of important edges in the predicted graphs. An edge is defined as important if not having it as part of the graph causes a decrease in the model's confidence for the target stance. We first fine-tune a RoBERTa model that given a \textit{(belief, argument, graph)} triple, predicts the probability of the target stance. Next, we remove one edge at a time from the corresponding graph and query the same model with the belief, argument and the graph but with the edge removed. If we observe a drop in the model's confidence for the target stance, the edge is considered important.

\section{Models}
\label{sec:model}

Following prior work on explanation generation \cite{rajani2019explain}, we experiment with two broad families of models -- (1) \textbf{Reasoning} (First-Graph-Then-Stance) models that first predict the explanation graph by conditioning on the belief and the argument. Then it augments the belief and the argument with the generated graph to predict the stance, (2) \textbf{Rationalizing} (First-Stance-Then-Graph) models that first predict the stance, followed by generating graphs as post-hoc explanations. In both these types of models, the stance prediction happens through a fine-tuned RoBERTa. For graph generation, we first propose a commonsense-augmented structured model (described next). We also experiment with state-of-the-art text generation models like BART and T5 that generate graphs as linearized strings. During training, edges in the graphs are ordered according to the depth-first-traversal (DFS) order of the nodes. See appendix for details on fine-tuning BART and T5 for graph generation.

\vspace{-4pt}

\paragraph{Commonsense-Augmented Structured Prediction Model:}

Next, as another baseline, we present a commonsense-augmented structured prediction model. As shown in Fig. \ref{fig:model}, it has the following four modules: (a) \textit{Internal Nodes Prediction:} It identifies the concepts (nodes) from the belief or the argument. We pose this task as a sequence-tagging problem where given a sequence of tokens from the belief and argument, each token is classified into one of the three classes \{B-N, I-N, O\} denoting the beginning, inside and outside of a node respectively. We build this module on top of a pre-trained RoBERTa \cite{liu2019roberta} by feeding in the concatenated belief and argument and having a standard 2-layer classifier at the top.
(b) \textit{External Commonsense Nodes Generation:} We build this module separately by fine-tuning a pre-trained BART \cite{lewis2019bart} model that conditions on the concatenated belief and argument and generates a sequence of commonsense concepts.
(c) \textit{Edge Prediction:} We pose this as a multi-way classification problem in which given a pair of nodes, the module has to classify the edge into one relation (or no edge). This module is conditioned on the node prediction module to enable learning edges between the set of chosen nodes only and also for optimizing both modules jointly. Specifically, given the set of node representations from the node module, we construct the edge representations for all possible edges which are then passed to a 2-layer classifier for prediction. To augment our model with external commonsense, we first fine-tune RoBERTa and the edge module on ConceptNet \cite{liu2004conceptnet}. 
(d) \textit{Enforcing Graph Constraints:} Our final loss sums the cross-entropy losses from the node and edge module. Following \citet{saha2020prover}, during inference, we ensure connectivity and acyclicity in the explanation graph through an Integer Linear Program. See appendix Sec. \ref{sp-model-appendix} for a more formal description of the model.\footnote{Our \data{} task also encourages future work based on other related structured models such as deep generative models for graphs~\cite{you2018graphrnn, simonovsky2018graphvae, grover2019graphite, liao2019efficient, shi2020graphaf}, but with adaptation to the unique challenges involved in our task, e.g., learning a good representation of the context using some pre-trained language model, identifying the internal nodes, generating the external/commonsense nodes and inferring the relations between nodes.}

\section{Experiments and Analysis}
\label{sec:expt}
In Table \ref{tab:model_test}, we compare our Reasoning-SP (RE-SP) model that generates graphs using the structured model with Rationalizing-BART/T5 (RA-BART/T5) and Reasoning-BART/T5 (RE-BART/T5) models that generate graphs using BART/T5. Besides our automatic metrics, the last column shows human evaluation of semantic correctness of graphs. Below, we summarize our key findings.

\paragraph{SP vs BART/T5:} BART and T5, used out-of-the-box, fail to generate a high percentage of structurally correct graphs (StCA) due to the lack of explicit constraints. Overall, RE-SP is the best performing model across all automatic metrics and human evaluation. It obtains a much higher StCA due to the constraints-enforcing ILP module and eventually a higher SeCA. Its superior performance is also reflected through the other metrics (G-BS, GED, and EA). See appendix for more analysis (like performance at varying reasoning depths, structures and the effect of edge ordering on BART/T5). 

\paragraph{Explanation Impact (RA vs RE): }RA models predict the stance first without the graph, while the RE models predict the stance conditioned on the generated graph. RE models' drop in SA points to their overall limitations in generating helpful explanations. In fact, conditioning on such graphs makes the model less confident of its stance predictions. 

\begin{figure}[t]
	\centering
    \includegraphics[clip, width=0.91\columnwidth]{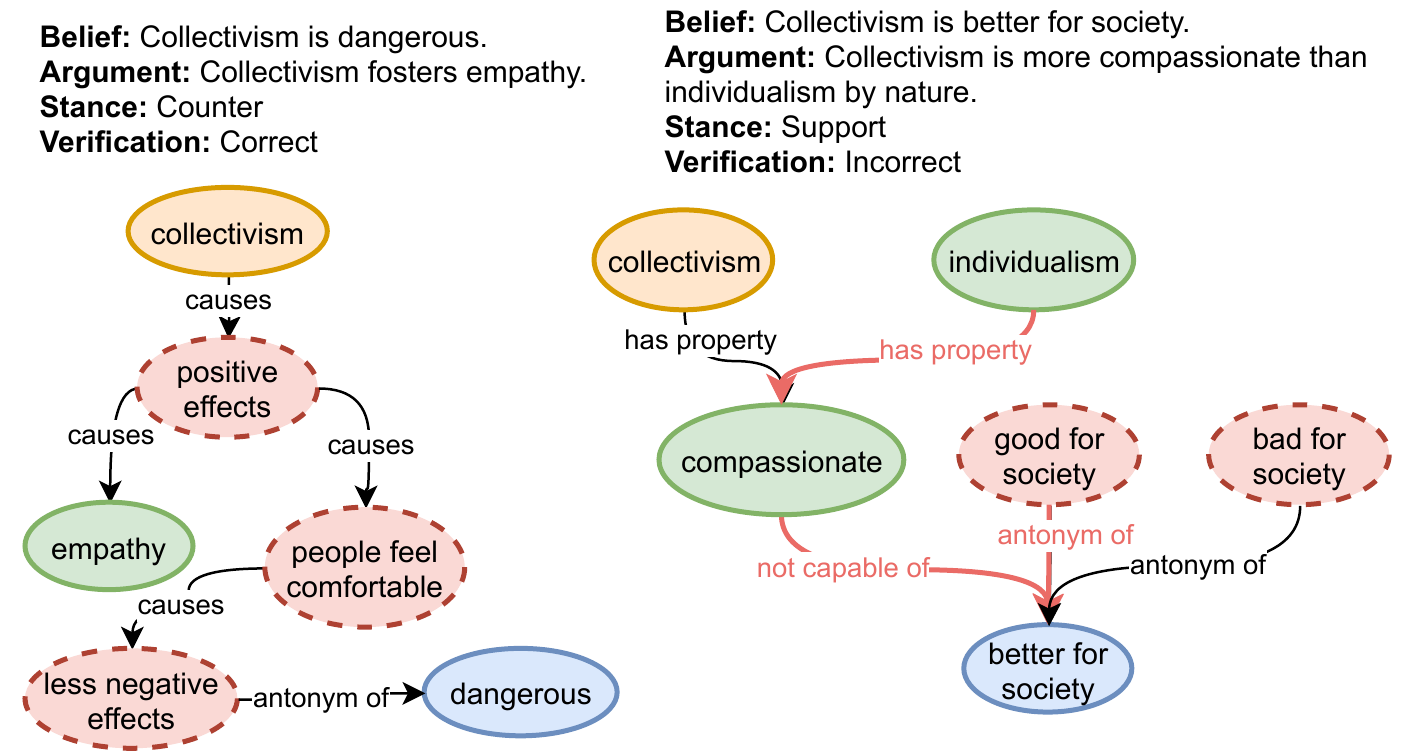}
    \vspace{-5pt}
    \caption{\label{fig:structured_example} Predicted graphs from the RE-SP model.  The first graph is correct, while the second one is not.} 
    \vspace{-10pt}    
\end{figure}

\paragraph{Metrics' Upper Bound:} While the stance accuracy (SA) is sufficiently high for all models, they obtain a significantly low semantic correctness accuracy (SeCA) for graphs (between 10-20\%). To obtain an upper bound on our metrics (last row), we treat ground-truth graphs as predictions and find that they not only aid in stance prediction (SA increases from 87\% to 91\%) but also obtain a high 83\% SeCA. Given the large gap (>60\%) between human and model performance, we hope our dataset will encourage future work on better model development for explanation graph generation.

\paragraph{Human-Metric Correlation for SeCA: }While we develop an initial automatic metric for SeCA, it still is a challenging problem and hence human evaluation for the same is necessary. In order to show human-metric correlation for SeCA, we perform human evaluation (using the exact mechanism of human-written graph verification, discussed in Sec. \ref{sec:graph_collection}) of all structurally-correct generated graphs. Encouragingly, we find that our model-based metric (SeCA column) correlates well with humans (last column), with RE-SP being the best model. The human verification labels match with the SeCA model's predictions 68\% of the time.

\paragraph{Analysis of Generated Graphs:} Fig. \ref{fig:structured_example} shows two randomly chosen graphs generated by RE-SP containing external commonsense nodes like ``positive effects'', ``good for society''. While the first graph is correct, the second graph chooses the wrong relations for certain edges (in red), thus pointing to its lack of commonsense. Overall, we find a large fraction of incorrect graphs contain incoherent facts or facts not adhering to human commonsense. Table \ref{tab:pred_stats} shows that RE-SP generates more nodes, edges, external nodes and non-linear structures, due to its individual components.

\section{Discussion and Future Work}

We show the promise of explanation graphs by considering the task of stance prediction as a motivating use-case because it is representative of many sentence-pair inference tasks (consider the belief as the premise, argument as the hypothesis and the support/counter labels as entailment/contradiction). We believe that our definition of explanation graphs (Sec. \ref{sec:task_definition}) is quite generic and should extend naturally to any NLU task, e.g., the internal nodes are concepts that are part of a context (context could mean premise-hypothesis for NLI, passage for sentiment classification, passage-question for QA, etc), the external nodes refer to concepts that are not part of the context, the edges are semantic relations between concepts, and the DAG-like constraints ensure the presence of explicit reasoning structures. Although we choose a pre-defined set of relations for our task that can adequately represent most commonsense facts, the relations can be updated/adapted for a different task. Given the potential of explanation graphs in improving the explainability of many reasoning tasks, we hope future work can further explore their applicability in different scenarios.

\begin{table}[t]
\small
    \centering
    \begin{tabular}{lrrrrrr}
    \toprule
         &  \#N & \#E & \#EN & D & \%NL & \%EN\\ \midrule
         RA-T5 & 4.3 & 3.3 & 0.3 & 3.3 & 3.5 & 24.8 \\
         RE-T5 & 4.4 & 3.4 & 0.3 & 3.3 & 7.6 & 28.8 \\
         RE-SP & 6.2 & 5.2 & 2.2 & 2.6 & 99.6 & 97.6 \\ \bottomrule
    \end{tabular}
    \vspace{-5pt}
    \caption{Statistics for the generated explanation graphs. NL = Non-Linear graphs, EN = External Nodes.}
\vspace{-10pt}
    \label{tab:pred_stats}
    
\end{table}

\section{Conclusion}
We proposed \data{}, a new \textit{generative} and \textit{structured} commonsense-reasoning task (and a benchmarking dataset) on explanation graph generation for stance prediction. Additionally, we proposed automatic evaluation metrics and an initial structured model for \data{}, demonstrating its difficulty in generating high-quality commonsense-augmented graphical explanations, and encouraging future work on better graph-based commonsense explanation generation.

\section*{Ethical Considerations}

We select crowdworkers from Amazon Mechanical Turk (AMT) who are located in the US and Australia with a HIT approval rate higher than 96\% and at least 1000 HITs approved. To ensure high data quality, we perform multiple on-boarding tests (details in the appendix) and manually verify a lot of the initial explanation graphs. We also provide personal feedback to a number of annotators. A total of 198 workers took part in our data collection and human verification process. We compensated annotators at the rate of \$12-15 per hour. The payments per HIT for each of our tasks are listed in the appendix. To estimate this, we first post small pilot studies to evaluate average
time of completion, and pay users accordingly. Annotators who annotated high-quality graphs were regularly  compensated with bonuses, throughout the duration of our data collection process. Also, our dataset mostly reflects the views of a set of English-speaking US annotators about some of the debate topics. However, for completeness, we collect both support and counter sides of the arguments. While some of the beliefs may span controversial topics, we as authors do not promote or stand with either side of the argument. Instead, we focus on the explainability aspect of these arguments through background commonsense knowledge.

\section*{Acknowledgements}
We thank the reviewers as well as Yejin Choi, Peter Clark, Peter Hase, Hyounghun Kim, and Jie Lei for their helpful feedback, and the annotators for their time and effort. This work was supported by DARPA MCS Grant N66001-19-2-4031, NSF-CAREER Award 1846185, Microsoft Investigator Fellowship, Munroe \& Rebecca Cobey Fellowship, and an NSF Graduate Research Fellowship. The views in this article are those of the authors and not the funding agency.

\bibliographystyle{acl_natbib}
\bibliography{emnlp2021}

\appendix
\section{Data Collection}

\begin{figure}[tbh!]
	\centering
    \includegraphics[clip, width=\columnwidth]{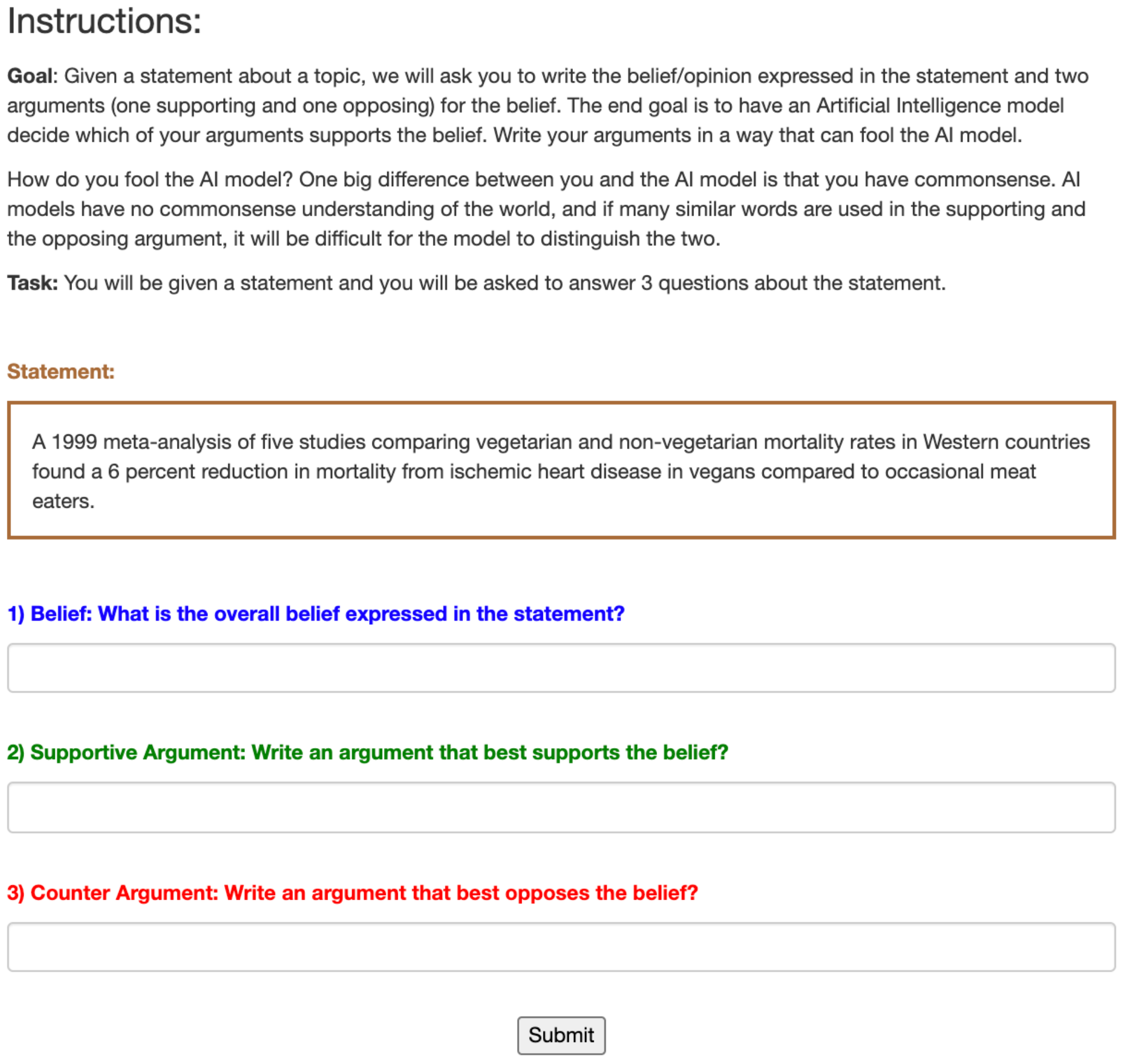}
     \caption{Interface showing the instructions for collecting belief and argument (support and counter) pairs on MTurk for the pre-HAMLET stage, given a prompt about one of the debate topics.\label{fig:instructions_stage1_prehamlet}} 
\end{figure}

\begin{figure}[tbh!]
	\centering
    \includegraphics[clip, width=\columnwidth]{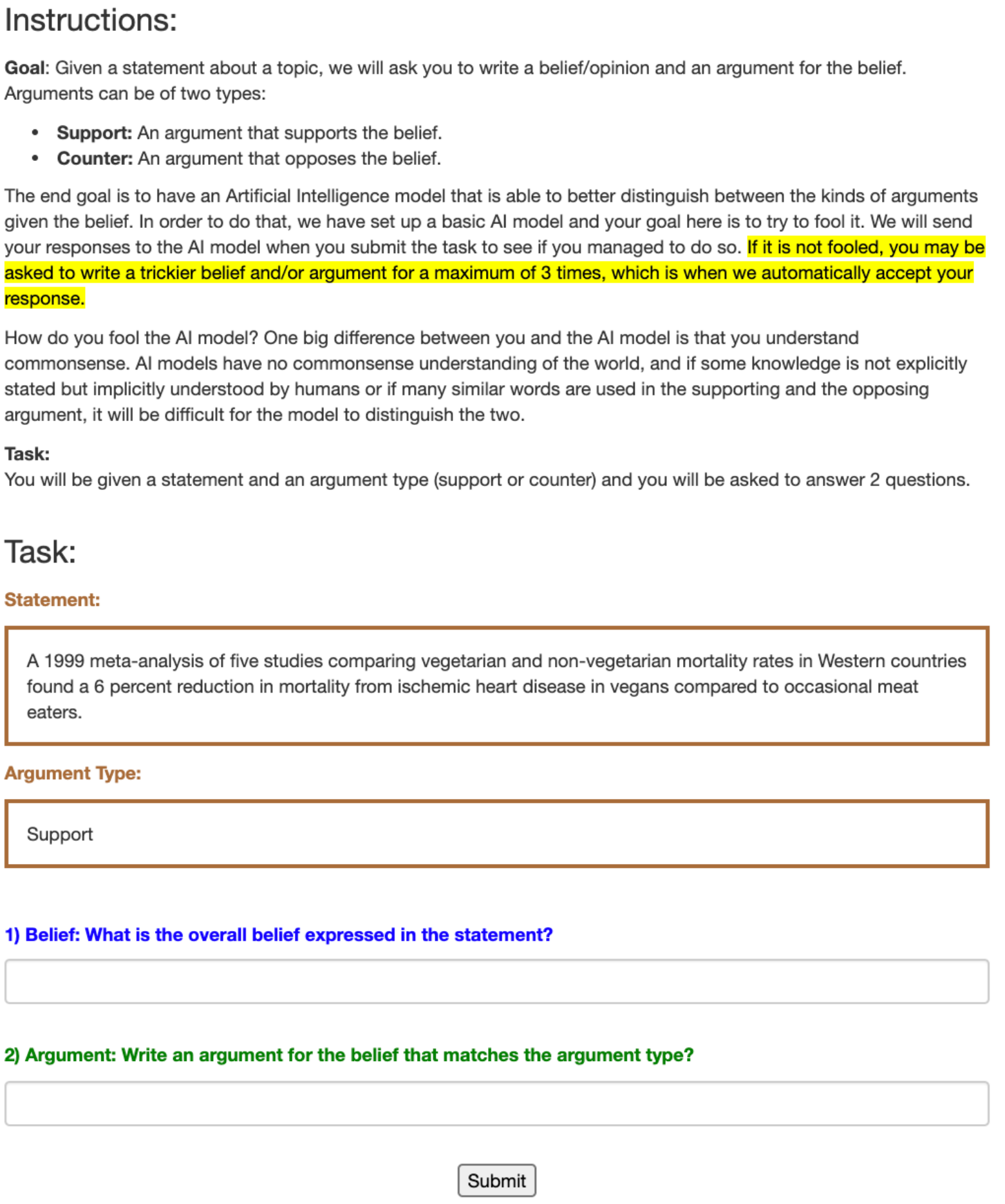}
     \caption{Interface showing the instructions for collecting belief and argument pairs on MTurk for the HAMLET stage, given a prompt about one of the debate topics and the target stance label (support or counter). \label{fig:instructions_stage1_hamlet}} 
\end{figure}

\subsection{Stage 1: (Belief, Argument, Stance) Collection}

Our stage 1 data collection consists of one pre-HAMLET round and three rounds of HAMLET \cite{nie2019adversarial}.

\paragraph{Pre-HAMLET:} The complete instructions for pre-HAMLET data collection is shown in Fig. \ref{fig:instructions_stage1_prehamlet}. Briefly, annotators write the belief expressed in the prompt along with a supporting and a counter argument. The beliefs and arguments are typically one-sentence long. We collect a total of 998 samples from randomly chosen 33 topics out of the 53 train topics with an average of 30 samples per topic. Note that we do not include the dev and test topics as part of the pre-HAMLET collection to ensure that the examples in these splits are sufficiently hard for the models. 

\begin{figure*}[t]
	\centering
    \includegraphics[clip, width=\textwidth]{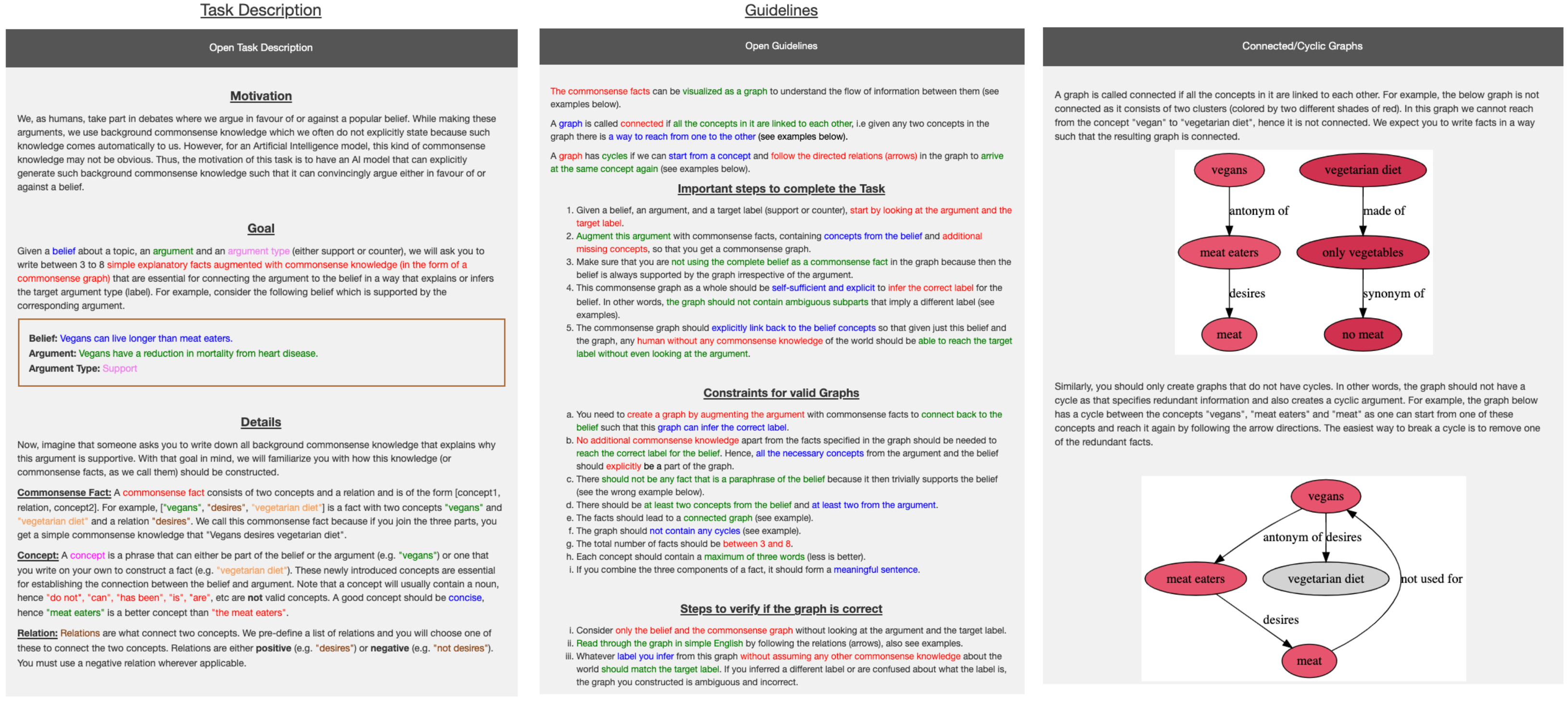}
     \caption{Instructions for commonsense explanation graph creation: We start by explaining the overall motivation and goal of this task, followed by the definitions of commonsense fact, concept, and relation. As part of the guidelines, we provide the detailed steps to perform this task and the list of structural constraints on the explanation graphs. We also remind the workers to verify their own graphs before submitting by following three basic steps of stance inference from the graphs. Since workers are required to fix their graphs if they are not connected DAGs, we also provide examples of disconnected and cyclic graphs.\label{fig:instructions_stage2_collection}} 
\end{figure*}

\begin{figure}[t]
	\centering
    \includegraphics[clip, width=\columnwidth]{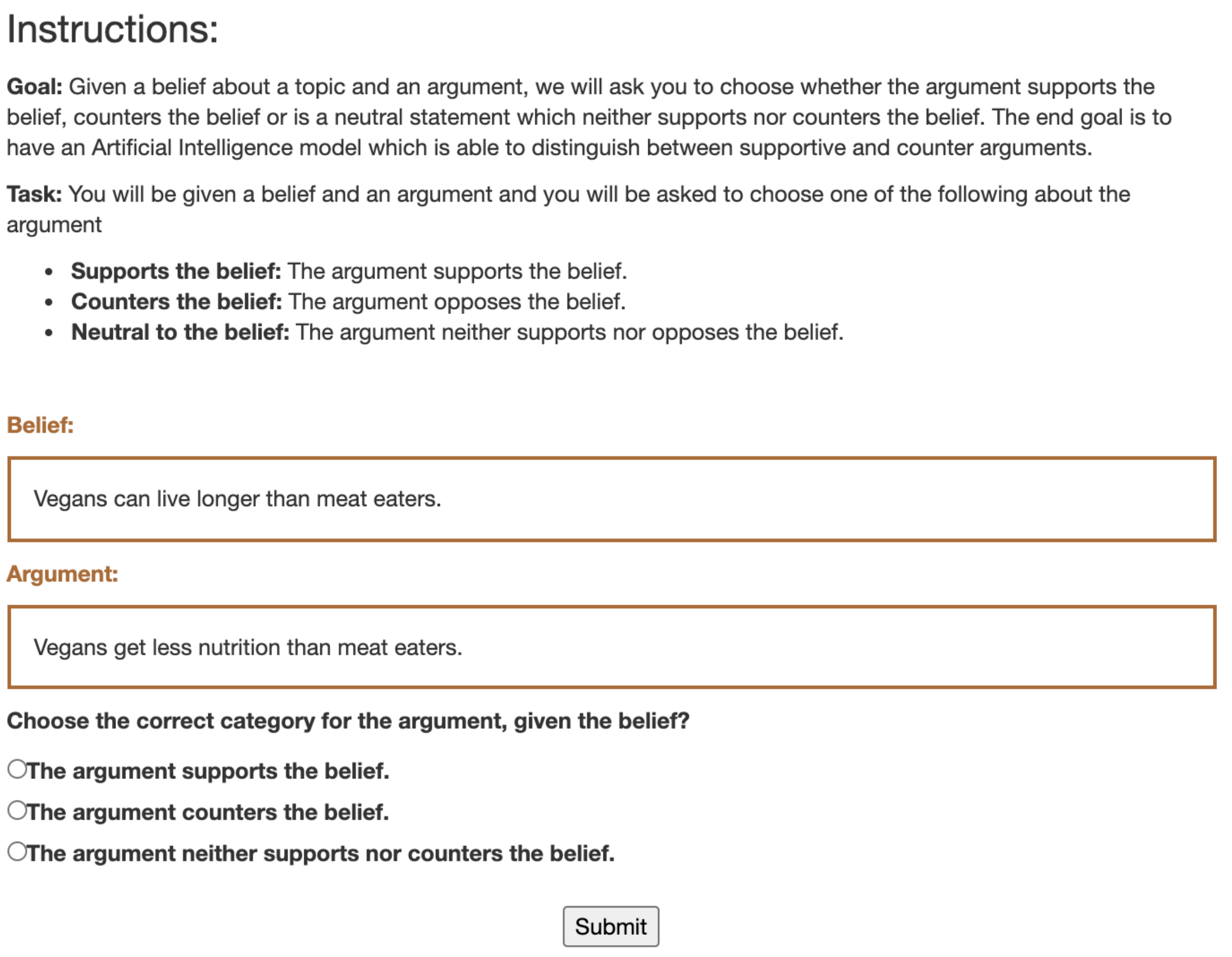}
     \caption{Interface showing the instructions for verifying the stance labels for belief and argument pairs on MTurk. We keep only those pairs which have majority stance label support or counter across five verifiers. \label{fig:instructions_stage1_verification}} 
\end{figure}

\begin{figure*}[t!]
	\centering
    \includegraphics[clip, width=\textwidth]{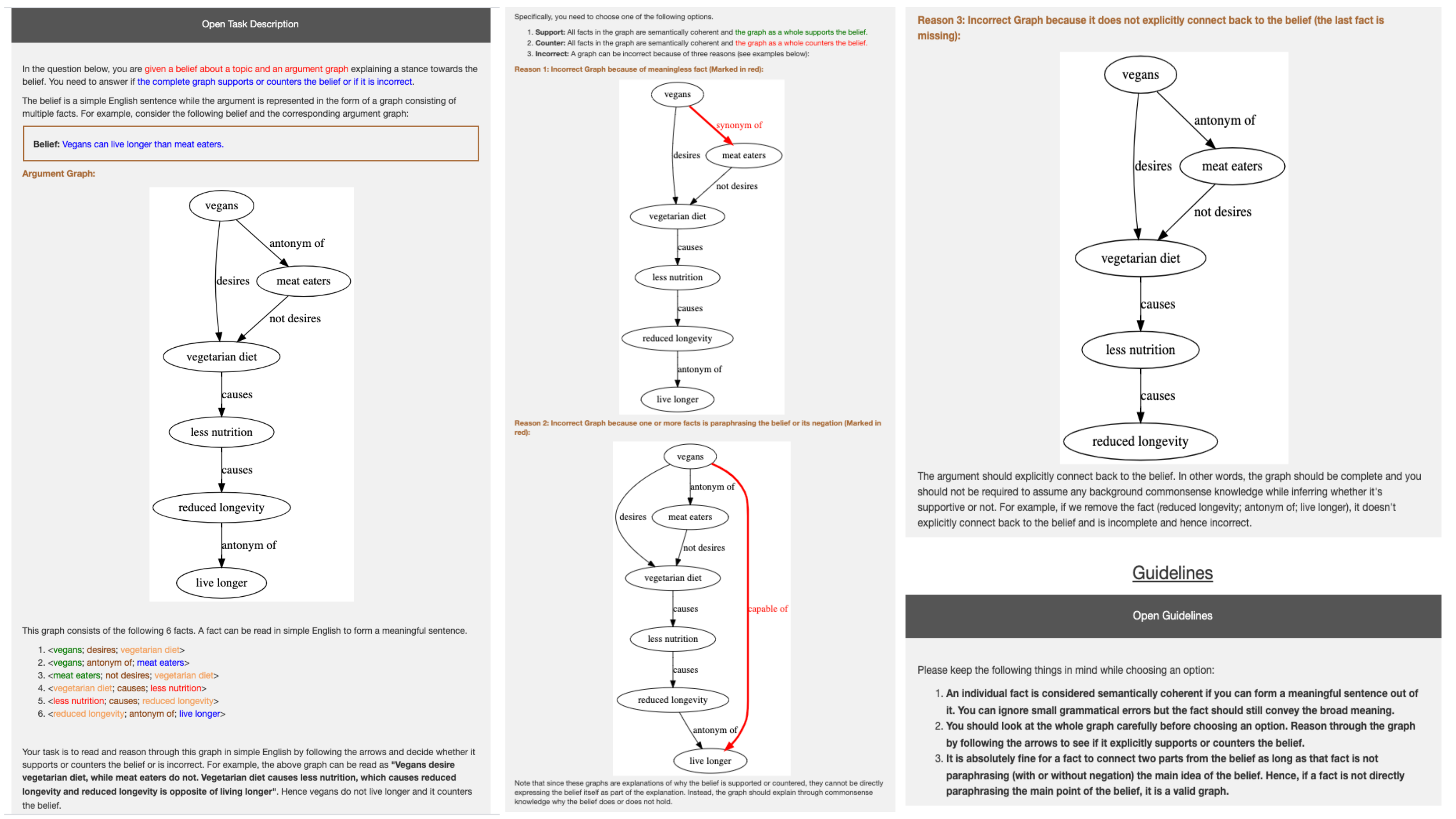}
     \caption{Instructions for commonsense graph verification: Explanation graphs are treated as augmented structured arguments for this task and hence referred to as argument graphs. Given a belief and the argument graph, workers are required to choose between incorrect, support and counter labels. We begin by visually explaining what an argument graph is, and also show examples of incorrect graphs. To ensure good inter-annotator agreement and that the semantically incorrect graphs are identified correctly, we also provide some general guidelines for performing this task. \label{fig:instructions_stage2_verification}} 
\end{figure*}

\paragraph{HAMLET:}

We follow the initial pre-HAMLET collection round with 3 rounds of HAMLET collection to reduce any annotation artifacts and most importantly, collect harder examples with implicit background knowledge. Fig. \ref{fig:instructions_stage1_hamlet} shows the instructions for the HAMLET rounds. At each round of HAMLET collection, we ask annotators to write (belief, argument) pairs in a way that a stance prediction model is fooled. In the first round, we start by fine-tuning a RoBERTa model \cite{liu2019roberta} on the pre-HAMLET data that given a (belief, argument) pair predicts the stance label. After each round, we divide the collected HAMLET data into train, dev and test splits based on their respective topics and update the RoBERTa model by training on the pre-HAMLET data and the train splits of the HAMLET rounds collected so far. We collect data in each round from the remaining 38 topics (20 train, 9 dev, 9 test) equally. In contrast to the pre-HAMLET round, here we also provide the target stance label along with the prompt and annotators are asked to write the belief and an argument that adhere to the target label. Once they construct a pair, in real-time, it is sent to the stance prediction model and if the model is able to predict the stance correctly, we prompt the annotators to rewrite either the belief or the argument. We provide annotators 3 tries in Round 1 and 4 tries in Round 2 and Round 3 to fool the model, following which we accept the final pair. Our HAMLET collection comprises of a total of 2170 samples with 892, 667 and 611 samples in rounds 1, 2, and 3 respectively.

\paragraph{Quality Control:} We apply the following mechanisms to control the quality of the collected data.

\begin{itemize}[nosep, wide=0pt, leftmargin=*, after=\strut]
    \item \textbf{Onboarding Test:} Each annotator is required to successfully pass an onboarding quiz before they can start writing belief and argument pairs. In this test, we evaluate their understanding of supportive and counter arguments by providing them with 10 (belief, argument) pairs and they are asked to choose if the argument supports or counters the belief.
    \item \textbf{Stance Label Verification:} We verify the stance labels of all the examples collected in pre-HAMLET and HAMLET rounds. This is particularly necessary for the HAMLET rounds where the annotators are constrained to fool the model and it is hard to create such samples and hence verification is required. Fig. \ref{fig:instructions_stage1_verification} shows the interface for our stance label verification, given the belief and the argument.
    For each (belief, argument) pair, we ask five annotators to choose the correct label between ``support'', ``counter'', and ``neutral''. We choose the majority label as the final label and keep only those examples that have majority labels either ``support'' or ``counter''.
\end{itemize}{}

\begin{figure}[tbh!]
	\centering
    \includegraphics[clip, width=\columnwidth]{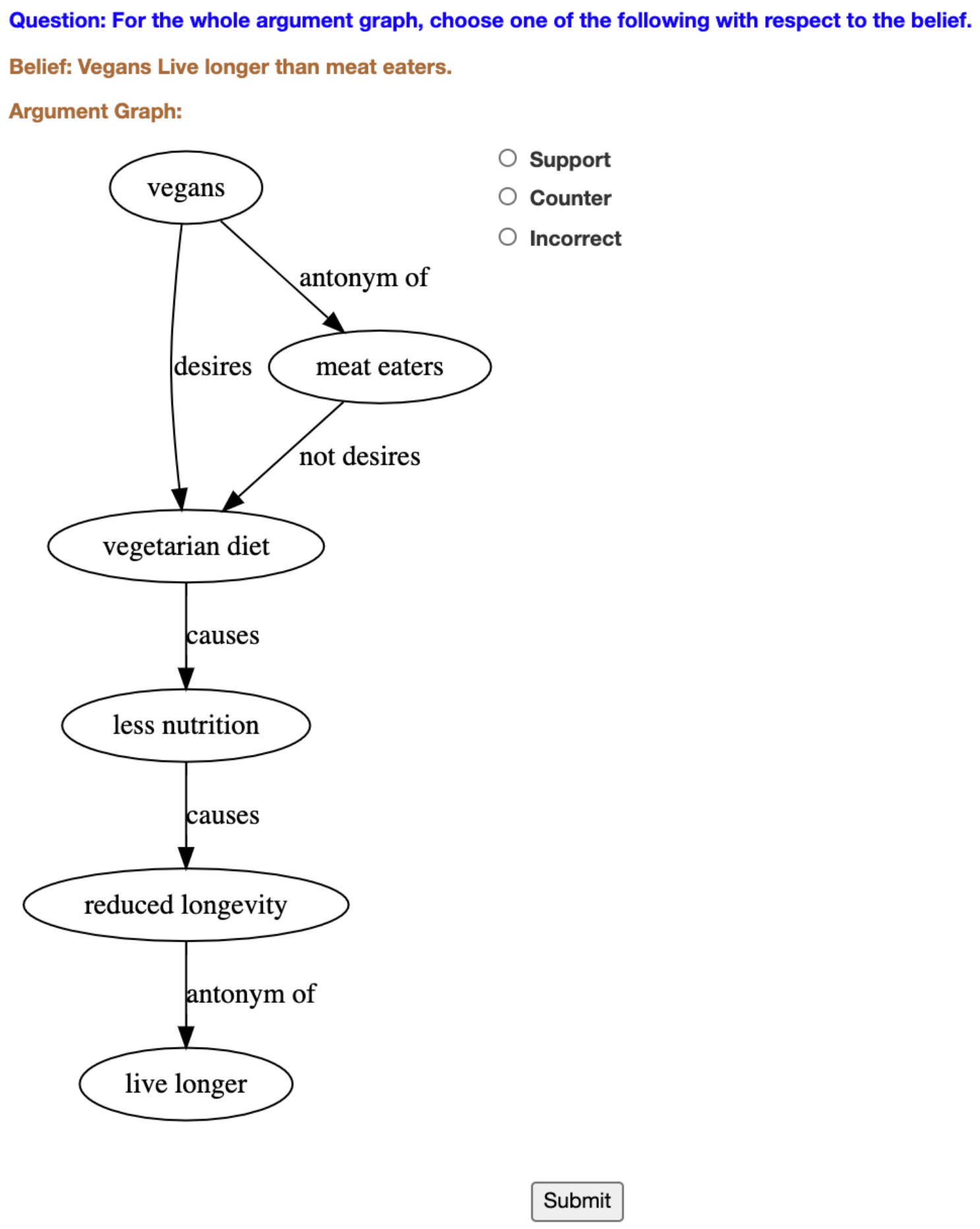}
     \caption{Interface for graph verification. \label{fig:interface_stage2_verification}} 
\end{figure}

\subsection{Stage 2: Commonsense Explanation Graph Collection}

\paragraph{Graph Creation:} Fig. \ref{fig:instructions_stage2_collection} shows the detailed instructions provided to the annotators for commonsense explanation graph creation. We start by explaining the overall motivation and the goal of our task, followed by the definitions of commonsense fact, concept, and relation. As part of the guidelines, we provide the detailed steps to perform this task and the list of structural constraints on the explanation graphs. We remind the workers to verify their own graphs before submitting, by following three basic steps of stance inference from the graphs. We also provide examples of disconnected and cyclic graphs to help them understand structurally incorrect graphs. 

\paragraph{Graph Verification:} In Fig. \ref{fig:instructions_stage2_verification}, we show the instructions provided for verifying the semantic correctness of our commonsense explanation graphs. In this stage, we refer to explanation graphs as argument graphs since our graphs are extended structured arguments. We provide annotators will only the belief and the argument graph, and ask them to choose between incorrect, support and counter labels. We also provide examples of semantically incorrect graphs. Fig. \ref{fig:interface_stage2_verification} shows the interface for graph verification. 

\begin{figure}[tbh!]
	\centering
    \includegraphics[clip, width=\columnwidth]{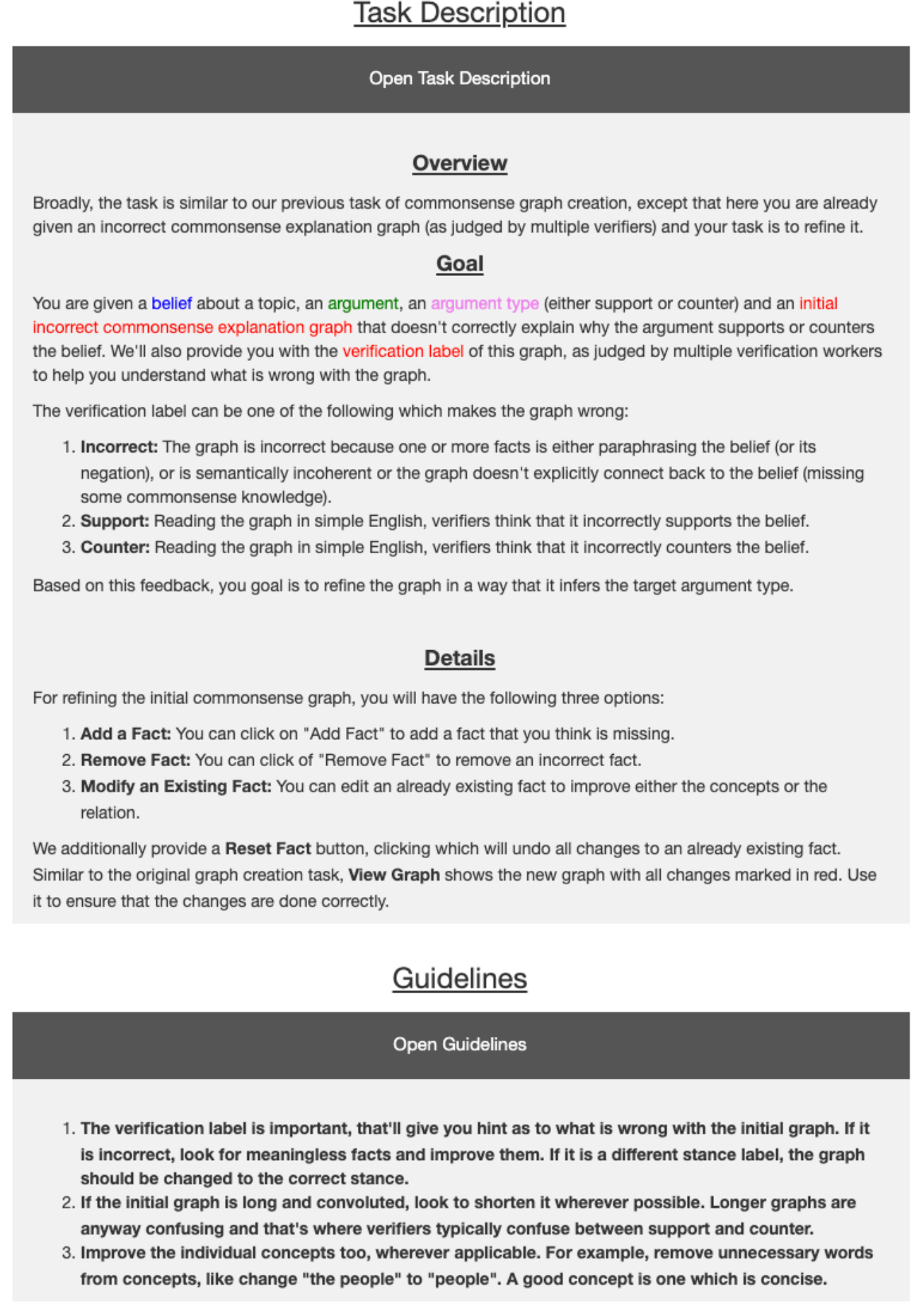}
     \caption{Instructions for graph refinement. \label{fig:instructions_stage2_refinement}} 
\end{figure}

\begin{figure}[tbh]
	\centering
    \includegraphics[width=\columnwidth]{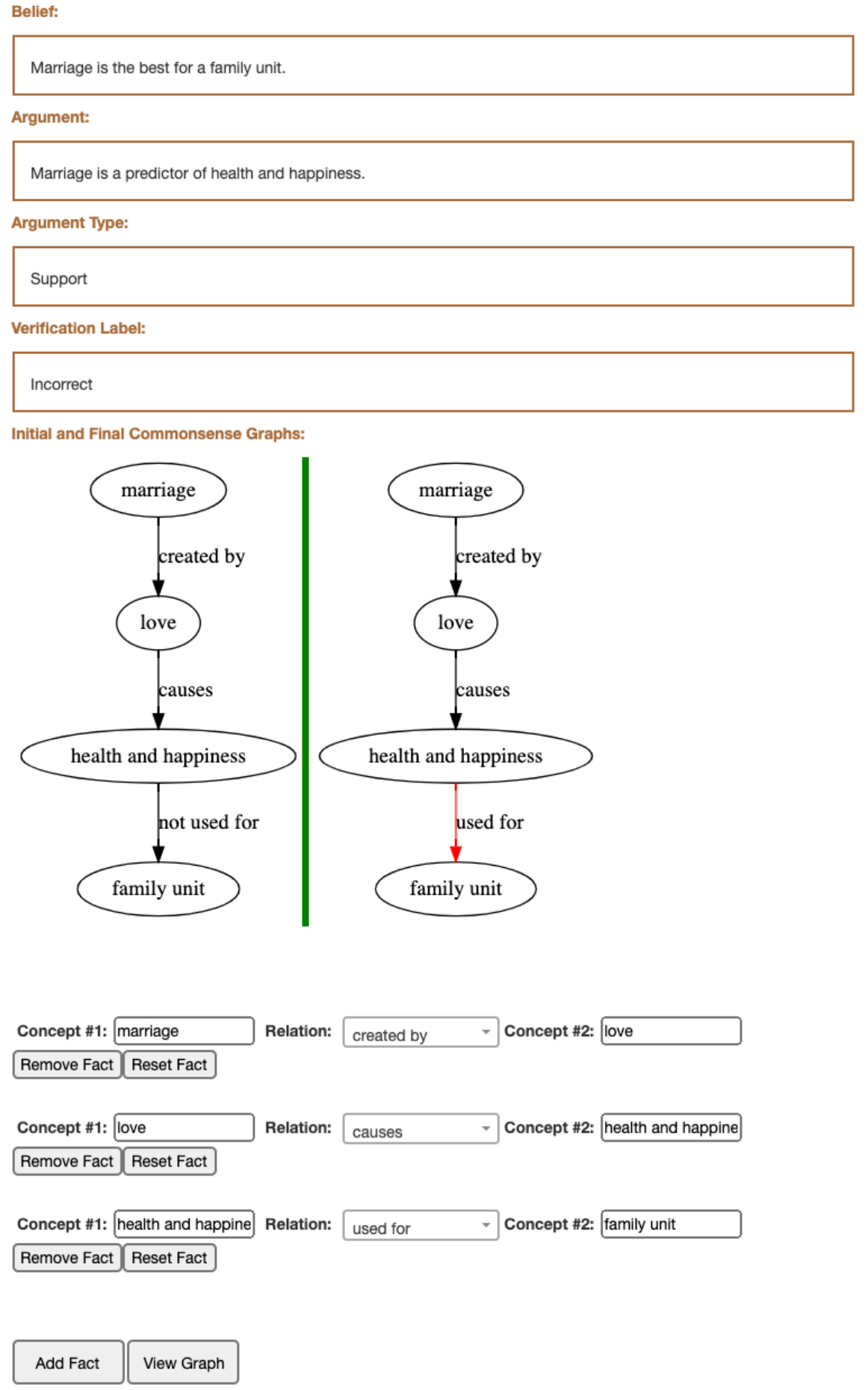}
     \caption{Interface for commonsense explanation graph refinement: Annotators are provided with the belief, argument, the stance label, the initial incorrect explanation graph and the majority verification label. They refine the graph by adding, removing or replacing facts and the changes to the initial graph are shown in red.}
     \label{fig:refinement_interface}
\end{figure}

\paragraph{Graph Refinement:} In Fig. \ref{fig:instructions_stage2_refinement}, we show the instructions of graph refinement in which we also provide some broad guidelines of how to refine the graphs. Our refinement interface is shown in Figure \ref{fig:refinement_interface}. They refine the initial graph by adding, removing or replacing facts and the ``View Graph'' button shows the updated graph, with the changes marked in red.

\paragraph{Quality Control:} Quality control of crowdsourced data is challenging, more so when the task involves creating graphs with associated constraints like acyclicity, connectivity, etc and then reasoning through the graph to infer the target label. Verifying these graphs for completeness, semantic coherence and non-triviality also requires understanding the overall motivation of the underlying task and hence is significantly more challenging than our Stage 1 stance label verification. In the light of these challenges, we employ carefully designed quality control mechanisms, which we believe will be helpful for similar graph collection tasks in the future.

\begin{figure*}[tbh!]
	\centering
    \includegraphics[clip, width=0.95\textwidth]{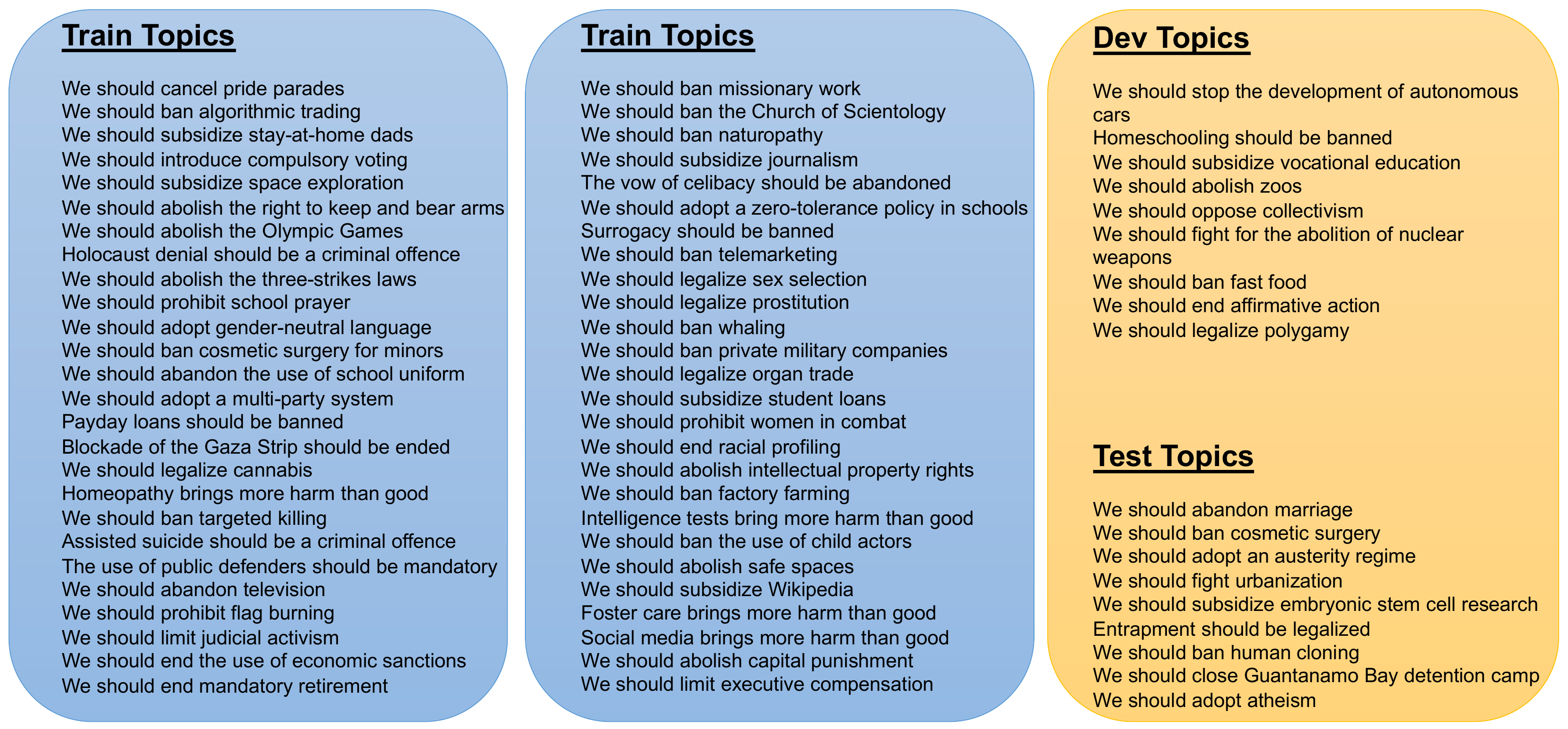}
     \caption{The complete list of debate topics used in our data collection process. \label{fig:topics}} 
\end{figure*}

\begin{figure}[tbh!]
	\centering
    \includegraphics[clip, width=0.65\columnwidth]{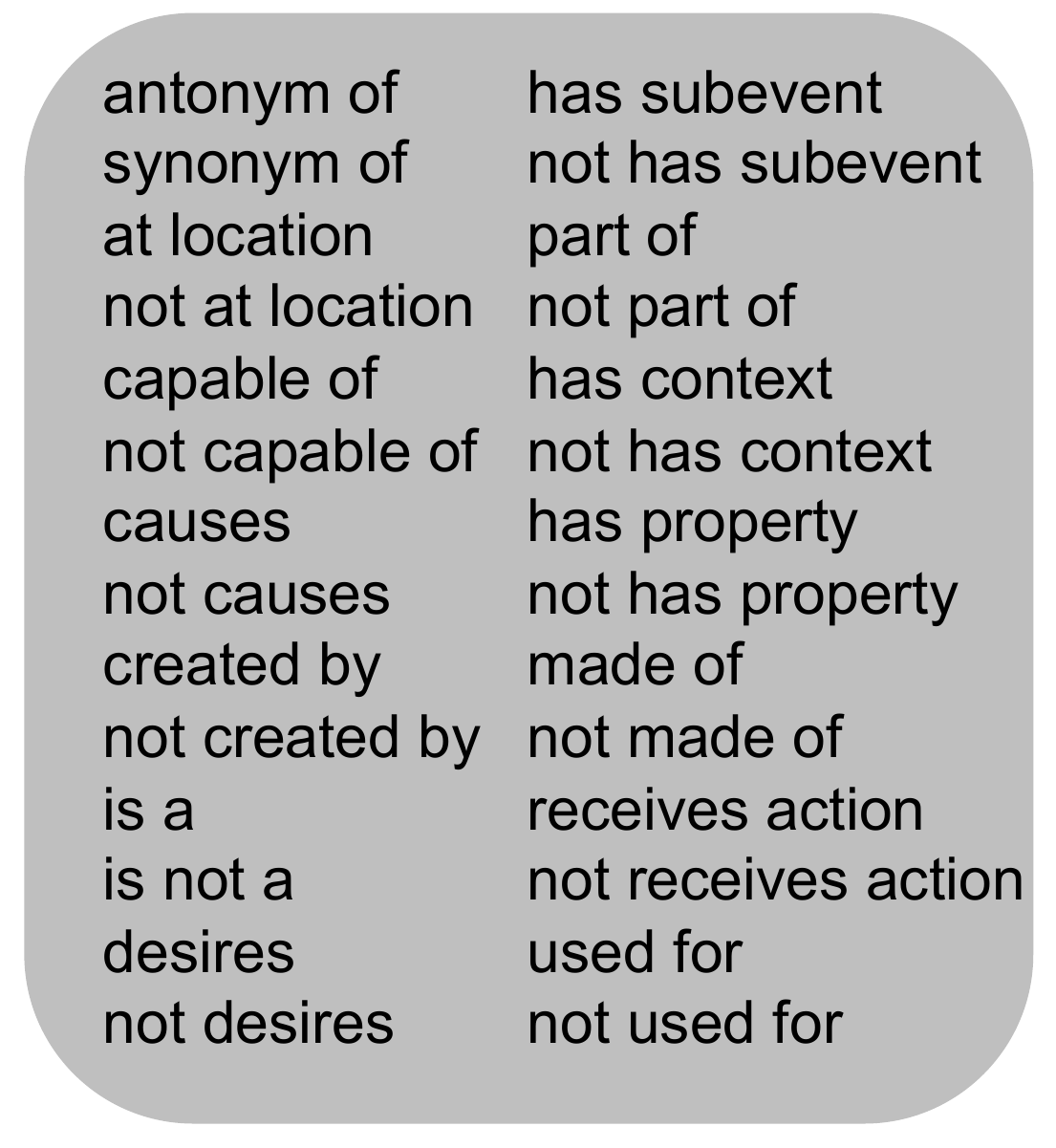}
     \caption{The complete list of commonsense relations used for our explanation graphs. \label{fig:relations}} 
\end{figure}

\begin{itemize}[itemsep=0pt, wide=0pt, leftmargin=*, after=\strut]
    \item \textbf{2-level Onboarding Test:} Since the three stages of graph creation, verification and refinement are closely tied to one another, we choose a single pool of annotators to perform all the graph-related tasks.
    We also prohibit annotators from verifying their own graphs. We  design a 2-level onboarding test where in the first level, we test the annotators' understanding of a commonsense fact because that is the basic building block of our graphs. Annotators are tested on 10 multiple choice questions, half of which require choosing the correct relation given the two concepts and another half require choosing the right pair of concepts, given the relation. Successful annotators from the first level qualify for the second level, where they are required to take two other tests. In one, we ask them to create a graph given a (belief, argument, stance) triple, whose quality we manually verify and in another, we ask them to verify the correctness of some already provided explanation graphs.

    \item \textbf{Intensive Training and Feedback:} We begin by providing detailed feedback and explanations of the correct answers from the onboarding tests to every qualified annotator. Every new annotator who starts creating graphs for the first time is initially requested to submit only a small number of graphs. We then verify these graphs manually and provide detailed feedback and suggest improvements wherever there are some incoherent facts in the graph or the graph is a trivial explanation or is incomplete. Over time, we find such personal feedback to be highly effective towards improving the quality of the graphs. 
    
    \item \textbf{High-performing annotators for Refinement:} While it is theoretically possible to run multiple iterations of graph verification and refinement, under most practical scenarios due to time and budget constraints, we want to ensure that a few rounds of refinement is enough to obtain a high percentage of correct graphs. Hence, we qualify only the high-performing annotators (whose graphs have been verified as correct the most) for our refinement task.
\end{itemize}{}

\begin{table}[tbh!]
\begin{tabular}{cc}
\toprule
\textbf{Task}         & \textbf{Pay/HIT (in cents)} \\ \midrule
Pre-HAMLET Collection & 25                          \\
HAMLET Collection     & 25                          \\
Stance Verification   & 5                           \\
Graph Creation      & 45                          \\
Graph Refinement      & 45                          \\ 
Graph Verification    & 10                          \\ \bottomrule 
\end{tabular}
\caption{Payment per HIT (in cents) for each of our tasks on MTurk (with additional bonuses). \label{fig:payment}} 
\end{table}

\begin{figure*}
\centering
\includegraphics[clip, width=\textwidth]{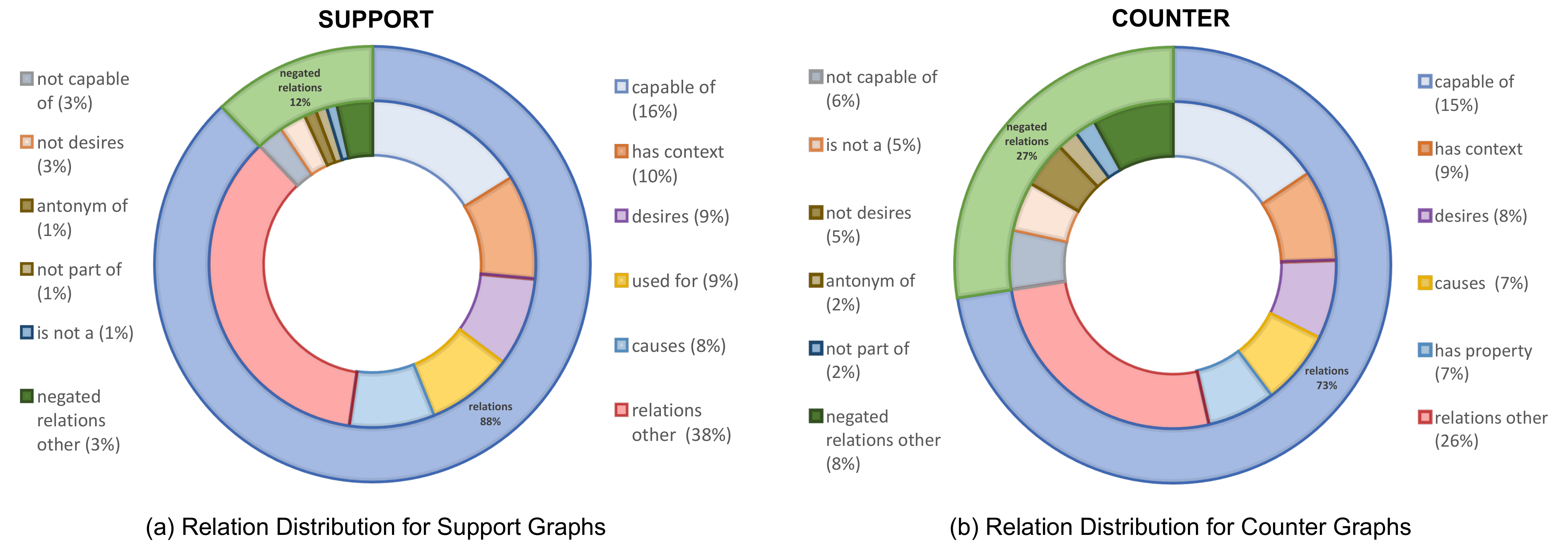}
      \caption{Relation percentages for Gold Graphs: Frequencies of occurrence of positive and negative relation for support and counter graph along with sub-classification into relation level statistics. \label{fig:plot_gold}}
\end{figure*}

\section{Data Analysis}

In Figure \ref{fig:topics}, we show the full list of debate topics used in our data collection process. The train split consists of 53 topics, while the dev and the test splits contain 9 topics each. Figure \ref{fig:relations} shows all the commonsense relations used for our explanation graph creation. We broadly choose the relation set from ConceptNet \cite{liu2004conceptnet}, while removing generic relations like ``related to'' and adding a negative counterpart for every positive relation to enable the composition of supportive and counter graphs.

Due to this, the relations used to construct the facts in our graphs can be divided into two categories -- with and without negations (``not capable of'' vs ``capable of''). We analyze the presence of these relations separately for the support and counter graphs. Fig. \ref{fig:plot_gold} illustrates that while non-negated relations are used more frequently in both kinds of graphs, they broadly follow a similar distribution of negated vs non-negated relations, demonstrating that the usage of a type of relation is not indicative of the stance label and actually depends on the specific context they are being used in. Interestingly, we also observe that the most frequently used relations in both stances are causal in nature (like ``capable of'', ``causes'', ``desires'', and their negative counterparts), which further supports our graphs as explanations.

\section{Models}

\subsection{Reasoning Model (First-Graph-Then-Stance)} Our first approach towards generating both stance and explanation graphs is through a reasoning model that first predicts the explanation graph by conditioning on the belief and the argument and then uses the generated graph, augmented with the belief and the argument, to predict the stance label. The explanation graph, in this case, provides additional commonsense knowledge and structure for the stance prediction task. For the BART \cite{lewis2019bart} or T5-based \cite{2020t5} graph prediction models, the input is the concatenated belief, argument (separated by separator) and the output is the explanation graph. We represent and predict graphs as linearized strings formed by concatenating the constituent edges. Since our explanation graphs are connected DAGs, during training, the edges are concatenated according to the depth-first-search (DFS) order of the nodes. In our experiments, we perform an empirical study showing that DFS marginally outperforms other edge orderings and is significantly better than a random ordering (see Results). Next, for the stance prediction model, we fine-tune a pre-trained sequence classification model, RoBERTa \cite{liu2019roberta}, which conditions on the concatenated belief, argument and the linearized graph to predict the stance label.\footnote{The stance prediction model can possibly be improved with better encoding of the explanation graph (e.g., through graph neural networks). We hope our challenging dataset encourages such model development as part of the future work by the community.}

\subsection{Rationalizing Model (First-Stance-Then-Graph)} Our second approach is via a rationalizing model which generates graphs as post-hoc explanations. Specifically, we first fine-tune a RoBERTa model to predict the stance label by conditioning on the belief and argument. The predicted labels are then concatenated with the belief and argument to fine-tune BART and T5 models for generating the explanation graph in a post-hoc manner. Similar to the reasoning models, graphs are represented as linearized strings according to the DFS order of the nodes.

\subsection{Commonsense-Augmented Structured Prediction Model}
\label{sp-model-appendix}

Our model consists of the following four components.

\paragraph{Internal Nodes Prediction:} It involves predicting the nodes which are either part of the belief or the argument. We build this module on top of RoBERTa \cite{liu2019roberta}, which takes in the concatenated belief and the argument (separated by a separator). The task is posed as a sequence-tagging problem, where given a sequence of tokens $s = (t_1, t_2, \hdots, t_n)$, each token is classified into one of the three classes, \{B-N, I-N, O\}, where B-N indicates the start token of the node, I-N denotes the intermediate node tokens and O denotes tokens which are not part of any node. For example, given a belief ``Factory farming should not be banned'', the gold sequence tag is \{B-N, I-N, O, O, O, B-N\}. Given the representation of each token from RoBERTa, we classify them into one of the three classes using two fully-connected layers with dropout. The module is trained using standard cross-entropy loss over all tokens.

\paragraph{External Commonsense Nodes Prediction:} For generating external commonsense nodes which are neither part of the belief nor the argument, we separately fine-tune a BART model.\footnote{We also experiment with T5, but find BART to perform better.} We construct samples, where the input is again the concatenated belief and the argument and the output is a comma-separated list of external nodes. For example, we construct samples like \textit{$X$ = Factory farming should not be banned <s> Factory farming feeds millions, $y$ = Food, Necessary}, where ``Food'' and ``Necessary'' are the commonsense nodes identified from the gold graph. The generated nodes from the BART model are fed to RoBERTa (from the previous module) and concatenated with the belief and the argument as part of the input, so as to have an unified model.

\paragraph{Edge Prediction:}
We model edge prediction as a multi-way classification problem over 29 classes (one class for each of our 28 relations and one for no edge, if no edge exists between the two nodes). Given the representation of each token from RoBERTa, we construct the representation of each node by mean-pooling over the representations of the constituent tokens. These node representations are used to construct the edge representations. Specifically, given two node representations $n_i$ and $n_j$ and the representation of a relation $r$, we construct the edge representation for that relation by concatenating the relation representation, the individual node representations along with their element-wise difference to capture the directionality of the edge.
Similar to the node module, the edge embeddings are also passed to a standard 2-layer classifier which predicts the probability of each edge belonging to any one of the classes. The module is trained with cross-entropy loss over all edges. Our final loss is the summation of the node loss and the edge loss. Given that our training data is not sufficient to learn commonsense relations between concepts from scratch, we initially fine-tune the RoBERTa pre-trained weights and the edge classifier on ConceptNet \cite{liu2004conceptnet} triples. Specifically, we consider facts like \textit{(man, capable of, eating)} from ConceptNet and create training data consisting of $X =$ \textit{man <s> eating} and $y =$ \textit{capable of} where \textit{<s>} is a separator used for separating the two concepts. We find that augmenting knowledge from ConceptNet improves the edge prediction capability of our model.

\paragraph{ILP Inference for Graph Constraints:}
\label{ilp}
Our inference procedure operates in two steps. Note that our edge prediction module is conditioned on the node module which means that edges will be predicted between the chosen nodes only. Thus, predicting edges requires predicting the nodes first. Once we obtain the internal and external nodes from their respective modules, in the second step, we predict the edge probabilities using the edge module. During edge inference, we want to enforce additional constraints such that the edges are predicted in a way that the final explanation graph is a connected DAG. Following prior work \cite{saha2020prover}, we achieve this through an Integer Linear Program (ILP) by maximizing a global score over the edge probabilities as described below.

\begin{table*}[tbh]
\small
\centering
\begin{tabular}{lrrrrrr}
\toprule
                    & SA$\uparrow$ & StCA$\uparrow$ & SeCA$\uparrow$ & G-BS$\uparrow$ & GED$\downarrow$ & EA$\uparrow$
                    \\ \midrule
RE-BART & 73.9 & 20.8 & 12.5 & 15.6 & 0.85 & 13.0 \\
RA-BART & \textbf{86.2} & 21.6 & 11.0 & 16.1 & 0.85 & 10.3 \\ 
RE-T5 & 70.8 & 29.6 & 12.2 & 22.8 & 0.79 & 18.0 \\
RA-T5 & \textbf{86.2} & 35.4 & 15.5 & 27.7 & 0.75 & 19.8 \\
RE-SP & 72.3 & \textbf{62.3} & \textbf{18.5} & \textbf{47.0} & \textbf{0.62} & \textbf{27.1} \\
\bottomrule  

\end{tabular}
\caption{Results of our Rationalizing and Reasoning models with structured, BART and T5 variants across all metrics on \data{} dev set.}
\label{tab:model_dev}
\end{table*}

\begin{table*}[tbh]
\small
\centering
\begin{tabular}{lrrrrrr}
\toprule
                     & SA$\uparrow$ & StCA$\uparrow$ & SeCA$\uparrow$ & G-BS$\uparrow$ & GED$\downarrow$ & EA$\uparrow$                \\ \midrule
Random & 70.3 & 9.3 & 4.0 & 7.1 & 0.93 & 5.6 \\
Topological & 69.6 & 27.6 & 11.0 & 21.2 & 0.81 & 16.1 \\
BFS & 70.3 & 28.9 & 10.2 & 22.4 & 0.80 & 16.1 \\
DFS & \textbf{70.8} & \textbf{29.6} & \textbf{12.2} & \textbf{22.8} & \textbf{0.79} &  \textbf{18.0}\\
\bottomrule                                             
\end{tabular}
\caption{Effect of edge ordering on Reasoning-T5 model. Having a random ordering leads to a significant drop in performance. Between fixed orderings, DFS performs better than BFS and Topological ordering.}
\label{tab:ordering}
\end{table*}

Checking for graph connectivity can be reduced to solving a max-flow problem in an augmented graph. Specifically, to ensure connectivity in an explanation graph $\mathcal{G} = (\mathcal{N}, \mathcal{E}$), we first define an augmented graph $\mathcal{G}_{aug} = (\mathcal{N}_{aug}, \mathcal{E}_{aug})$ with two additional nodes $s_o$ and $s_i$ representing a source node and a sink node respectively. We further add an edge from the source $s_o$ to any one of the nodes $n$ in $\mathcal{G}$ and from all nodes in $\mathcal{G}$ to the sink $s_i$. Now, for a graph to be connected, there should be a maximum total flow of $|\mathcal{N}|$ from $s_o$ to $s_i$.

In the reduced maximum-flow formulation \cite{leighton1999multicommodity} in $\mathcal{G}_{aug}$, we define a capacity variable $c_{(m,n)}$ for each edge, $m \rightarrow n$ in $G_{aug}$, as follows.
\begin{eqnarray}
c_{(s_o,x)} = |\mathcal{N}| \text{ and } c_{(x, s_o)} = 0 \nonumber \\
\forall n \in \mathcal{N}, c_{(n,s_i)} = 1 \text{ and } c_{(s_i,n)} = 0 \nonumber
\end{eqnarray}

Our final optimization problem is as follows. From our edge module, we obtain $e_{(m,n,r)}$, the probability that an edge $m \rightarrow n$ has the relation $r$. Additionally, we also obtain $e_{(m,n,-)}$, the probability that no edge exists between the nodes $m$ and $n$. Given these probabilities, we define binary optimization variables $\phi_{(m,n)}$, where 1 means that an exists between the nodes $m$ and $n$, while 0 means no such edge exists. Our final optimization function is:

\begin{equation}
\begin{split}
    \underset{\phi_{(m,n)}, f_{(m,n)}}{argmax}
    \sum_{m,n, m\neq n} (\phi_{(m,n)} {\max_r} (e_{(m,n,r)}) + \\ (1-\phi_{(m,n)})(e_{(m,n,-)})) \nonumber
\end{split}
\end{equation}
subject to constraints: 
\begin{eqnarray}
\small
    \forall m,n \in \mathcal{N}_{aug}, 0 \leq f_{(m,n)} \leq c_{(m,n)} \label{eq:const5}\\
    \forall n \in \mathcal{N}_{aug}, \sum_{m:(m,n)\in \mathcal{E}_{aug}} f_{(m,n)} \nonumber \\ =  \sum_{o:(n,o)\in \mathcal{E}_{aug}} f_{(n,o)}\label{eq:const7}\\
    f_{(s_o, x)} = |\mathcal{N}| \label{eq:const8}
\end{eqnarray}

Equations \ref{eq:const5} and \ref{eq:const7} define the flow constraints which state that flow for each edge is bounded by its capacity and that the total flow at each node is conserved. Finally, Equation \ref{eq:const8} ensures connectivity in the explanation graph, by enforcing the total flow to be $|\mathcal{N}|$. When an edge exists, we choose the relation $r$ with the maximum probability.

\begin{table*}[t]
\small
\centering
\begin{tabular}{lrrrrrr}
\toprule
                     & SA$\uparrow$ & StCA$\uparrow$ & SeCA$\uparrow$ & G-BS$\uparrow$ & GED$\downarrow$ & EA$\uparrow$                  \\ \midrule
Low (1-3) & 73.1 & 63.1 & 18.4 & 45.4 & 0.66 & 26.3 \\
Medium (4-5) & 74.1 & 64.8 & 19.1 & 50.6 & 0.65 & 29.7 \\
High (6-8) & 63.0 & 50.0 & 17.4 & 38.9 & 0.73 & 20.8 \\
\bottomrule  

\end{tabular}
\caption{Comparison of Reasoning-SP model on the subset of examples in \data{} dev set with varying reasoning depths (low, medium, high). Performance on the graph-related metrics drop significantly at higher depth.}
\label{tab:depth}
\end{table*}

\begin{table*}[tbh]
\small
\centering
\begin{tabular}{lrrrrrrr}
\toprule
                     & SA$\uparrow$ & StCA$\uparrow$ & SeCA$\uparrow$ & G-BS$\uparrow$ & GED$\downarrow$ & EA$\uparrow$                            \\ \midrule
Linear & 72.1 & 63.3 & 23.9 & 47.7 & 0.66 & 28.5 \\
Non-linear & 72.7 & 61.0 & 11.6 & 45.4 & 0.68  & 25.2 \\\bottomrule  

\end{tabular}
\caption{Comparison of Reasoning-SP model on the subset of examples in \data{} dev set with linear vs non-linear graph structures. The semantic correctness accuracy (SeCA) of graphs drops significantly for non-linear graphs due to the complex reasoning process involved in such graphs.}
\label{tab:structure}
\end{table*}

\section{Experimental Setup}
We train all our models using the Hugging Face transformers library \cite{wolf2019huggingface}.\footnote{\url{https://github.com/huggingface/transformers}} For all RoBERTa-based models (including the commonsense-augmented structured model and the stance prediction models and our model-based metrics), we use RoBERTa-large \cite{liu2019roberta} with a batch size of $32$, an initial learning rate of $10^{-5}$ with linear decay, a weight decay of $0.1$ and a maximum sequence length of $128$ for training up to a maximum of $10$ epochs. As for BART \cite{lewis2019bart} and T5 \cite{2020t5}, we use their base models with a batch size of $8$, an initial learning rate of $3*10^{-5}$ and train for a maximum of $6$ epochs. The maximum input and output sequence lengths are set to $100$ and $150$ respectively. Graphs are generated from these models using standard beam search decoding with beam size of 4. Batch size and learning rate are manually tuned in the range $\{8, 16, 32\}$ and \{$10^{-5}$, $2*10^{-5}$, $3*10^{-5}$\} respectively and the best models are chosen based on our validation set performance. The random seed is chosen as $42$ in all our experiments. The total number of parameters of our structured model is similar to that of RoBERTa-large (355M). All of our models have an average runtime between 30 mins to 1 hour. The ILP inference is modeled using PuLP.\footnote{ \url{https://pypi.org/project/PuLP/}} All experiments are performed on one V100 Volta GPU.

\section{Additional Experiments and Analysis}

Table \ref{tab:model_dev} shows the results of all models on the \data{} dev set.

\subsection{Effect of Edge Ordering in BART/T5} 

In order to evaluate the effect of a particular edge ordering on BART and T5 fine-tuning for graph generation, we compare the performance of the Reasoning-T5 model with edges ordered according to (1) a random order, (2) Topological, (3) Breadth First Search (BFS), and (4) Depth First Search (DFS). From Table \ref{tab:ordering}, we observe that having a pre-defined ordering enables the model to learn the graph structure significantly better. This, however, is not surprising; due to the auto-regressive nature of these text generation models, an un-ordered edge set confuses the model and it is not able to learn the structural properties of graphs. We observe that the random model often generates cycles and hence has a significantly low percentage of structurally correct graphs. Having a fixed ordering also enables the model to learn an inductive bias towards generating graphs in a manner than can be read and reasoned through by humans. Owing to the slightly better performance of DFS, we conduct all our experiments with the same ordering.

\subsection{Analysis with Reasoning Depths} 

We refer to the depth of a graph as the reasoning depth involved in inferring the stance label. As part of ablation analysis, in Table \ref{tab:depth}, we analyze the performance of the Reasoning-T5 model on the subset of examples requiring varying depths of reasoning from low (depth $<=$ 3) to high (depth $>$ 5). Unsurprisingly, we find that our task of explanation graph generation becomes challenging at higher depth, as demonstrated by a drop in all graph-related metrics at depth $>=6$. This reveals the hardness of our task and encourages future work on better model development of explanation graph generation.

\subsection{Analysis with Reasoning Structures}

Our next ablation analyzes the effect of linear vs non-linear reasoning structures. We call a reasoning structure linear when the explanation graph contains a single chain of nodes. A non-linear reasoning structure adds complexity to the inference process and we validate this through our results in Table \ref{tab:structure}. Similar to the previous result, we observe that our task becomes challenging with non-linear structures as demonstrated by a significant drop in semantic correctness accuracy.

\begin{figure}[tbh!]
	\centering
    \includegraphics[clip, width=\columnwidth]{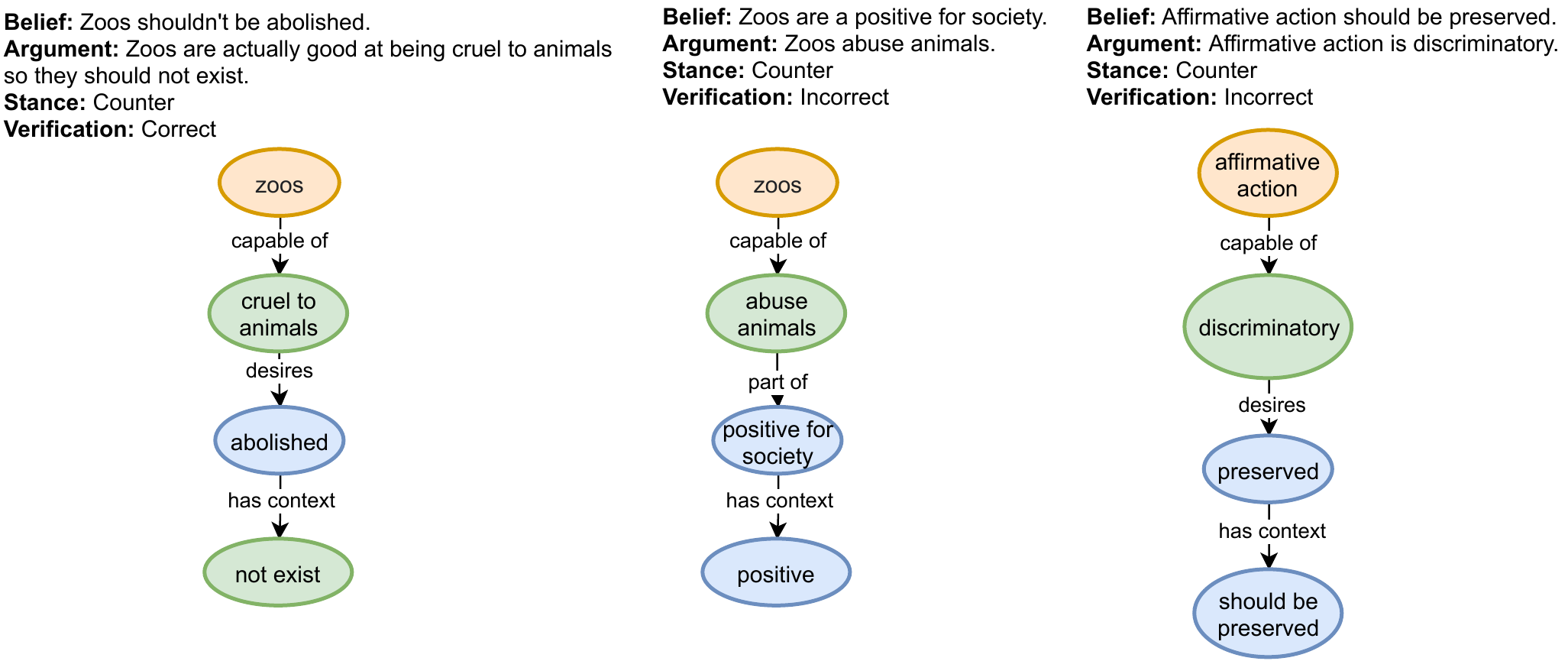}
     \caption{\label{fig:gen_t5}Examples of predicted graphs from the Reasoning-T5 model. The verification term stands for the outcome of human verification while stance refers to the gold label for the (belief, argument) pair.} 
\end{figure}

\subsection{Quantitative Analysis of Generated Explanation Graphs from RE-T5}
In order to gain a better understanding of the explanation graphs generated by our Reasoning-T5 model, we show sample explanation graphs generated by the model in Figure \ref{fig:gen_t5}. Unlike our RE-SP model, it typically generates linear chains with much fewer number of external commonsense nodes.

\section{Examples from \data{}}
\begin{figure}[tbh!]
	\centering
    \includegraphics[clip, width=\columnwidth]{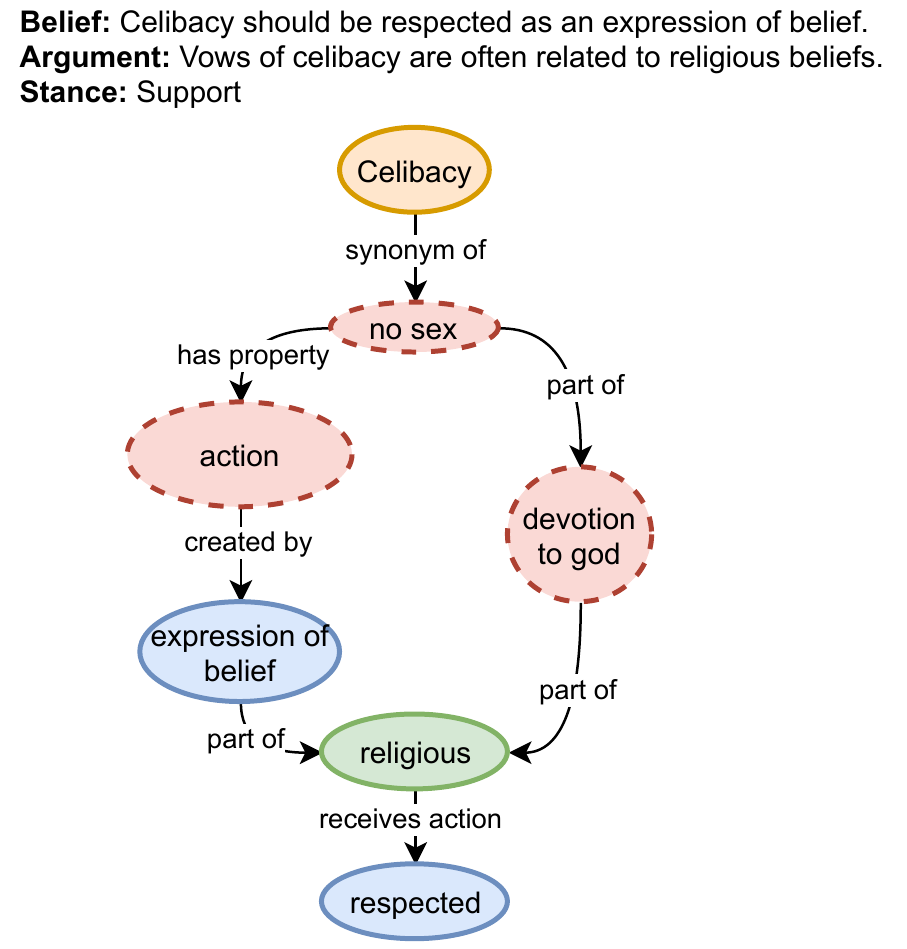}
     \caption{\label{fig:example3}Example 1} 
\end{figure}

\begin{figure}[tbh!]
	\centering
    \includegraphics[clip, width=\columnwidth]{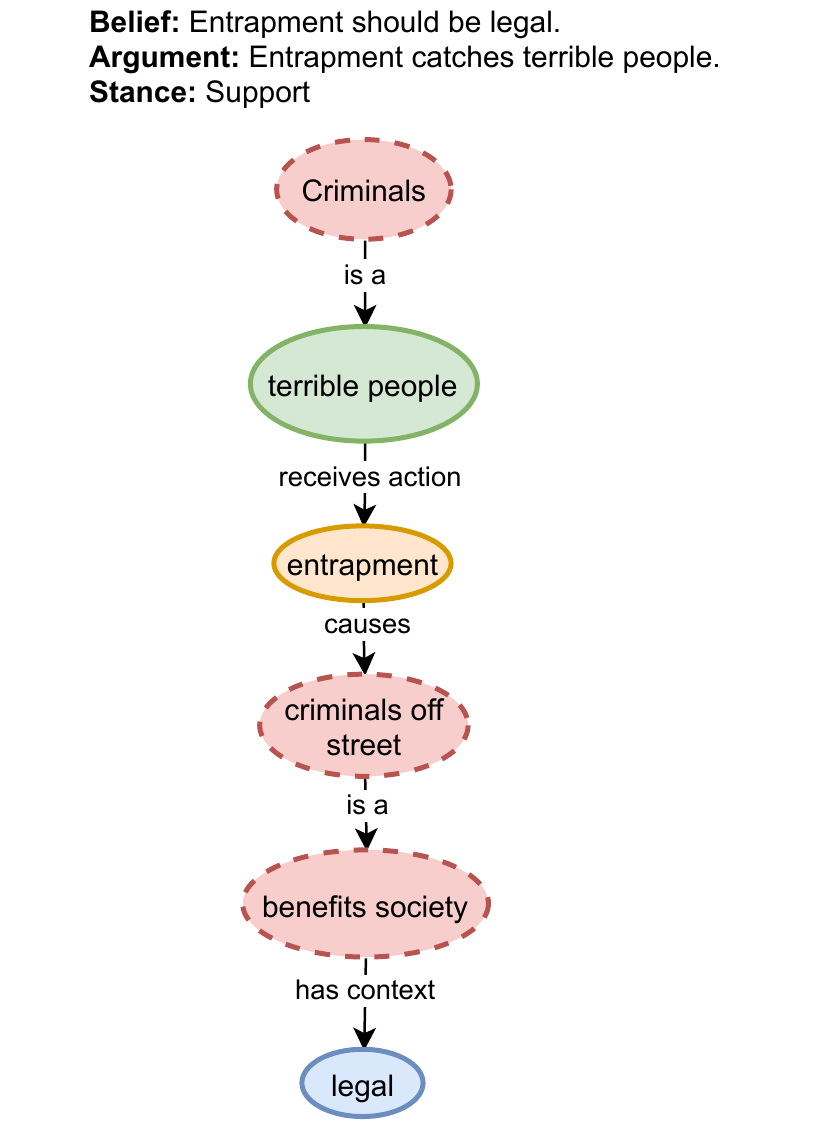}
     \caption{\label{fig:example4}Example 2} 
\end{figure}

\begin{figure}[tbh!]
	\centering
    \includegraphics[clip, width=\columnwidth]{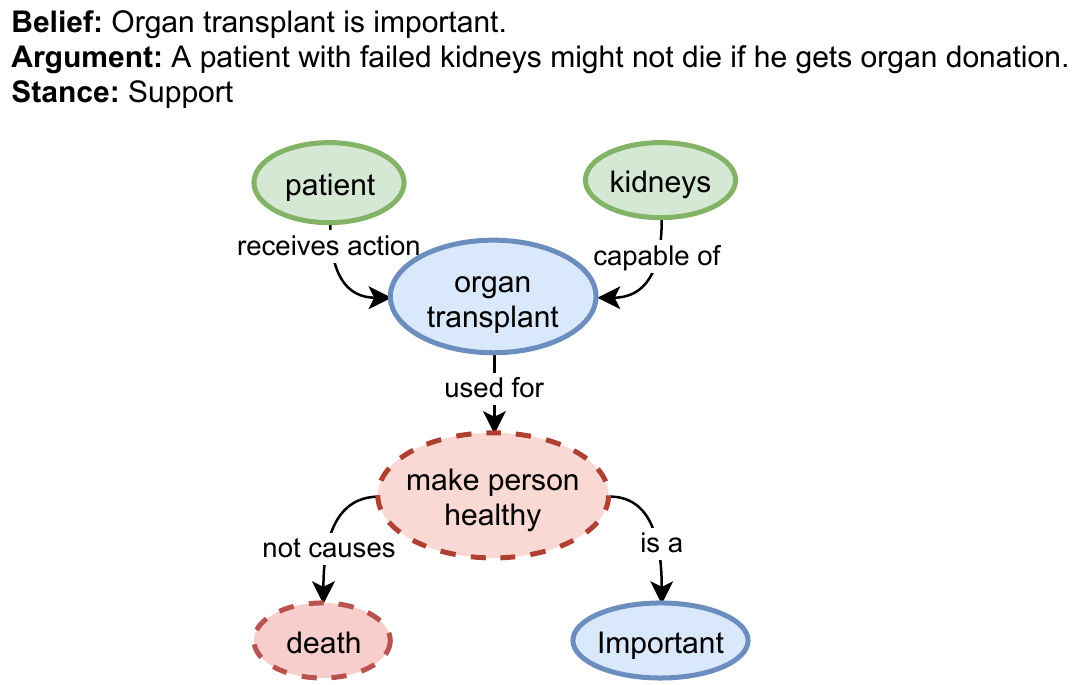}
     \caption{\label{fig:example5}Example 3} 
\end{figure}

\begin{figure}[tbh!]
	\centering
    \includegraphics[clip, width=\columnwidth]{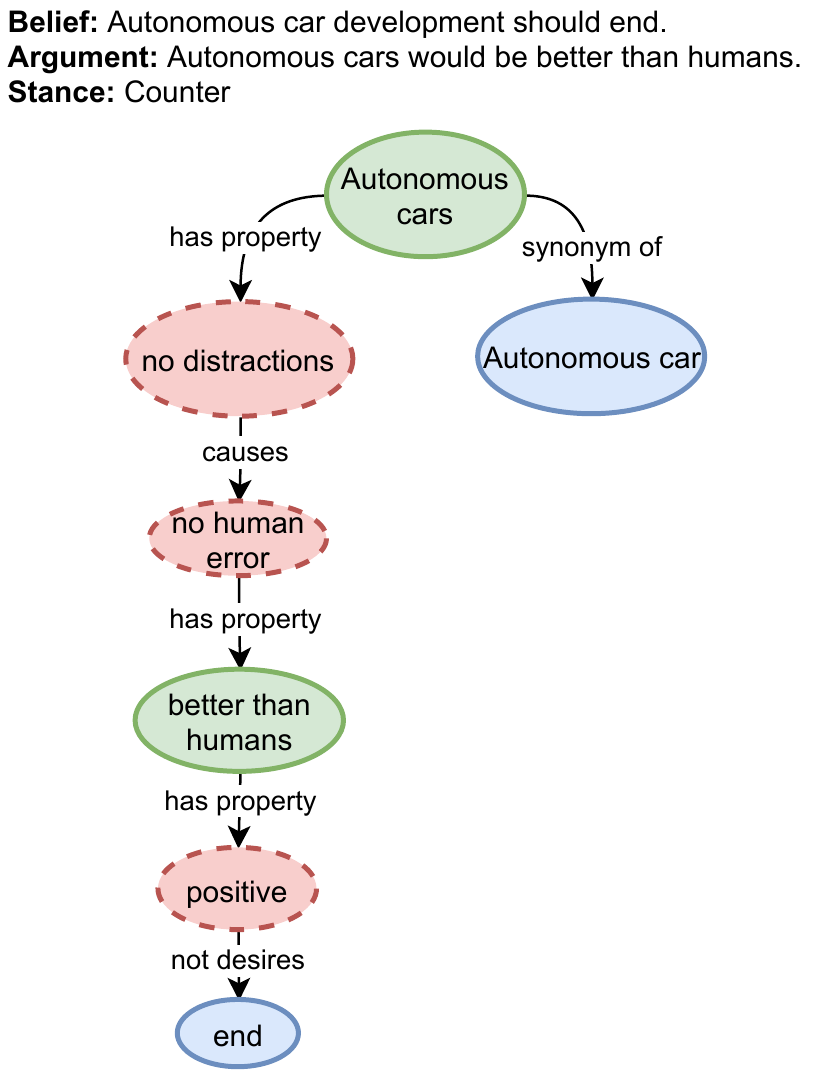}
     \caption{\label{fig:example6}Example 4} 
\end{figure}

\begin{figure}[tbh!]
	\centering
    \includegraphics[clip, width=\columnwidth]{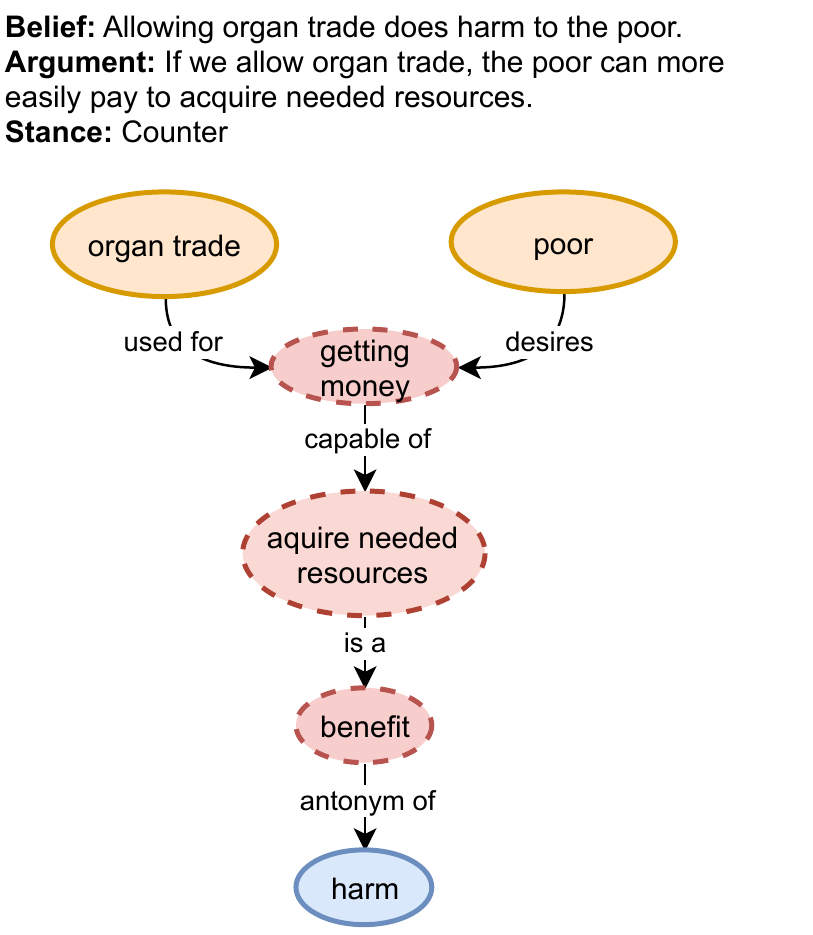}
     \caption{\label{fig:example7}Example 5} 
\end{figure}

\begin{figure}[tbh!]
	\centering
    \includegraphics[clip, width=\columnwidth]{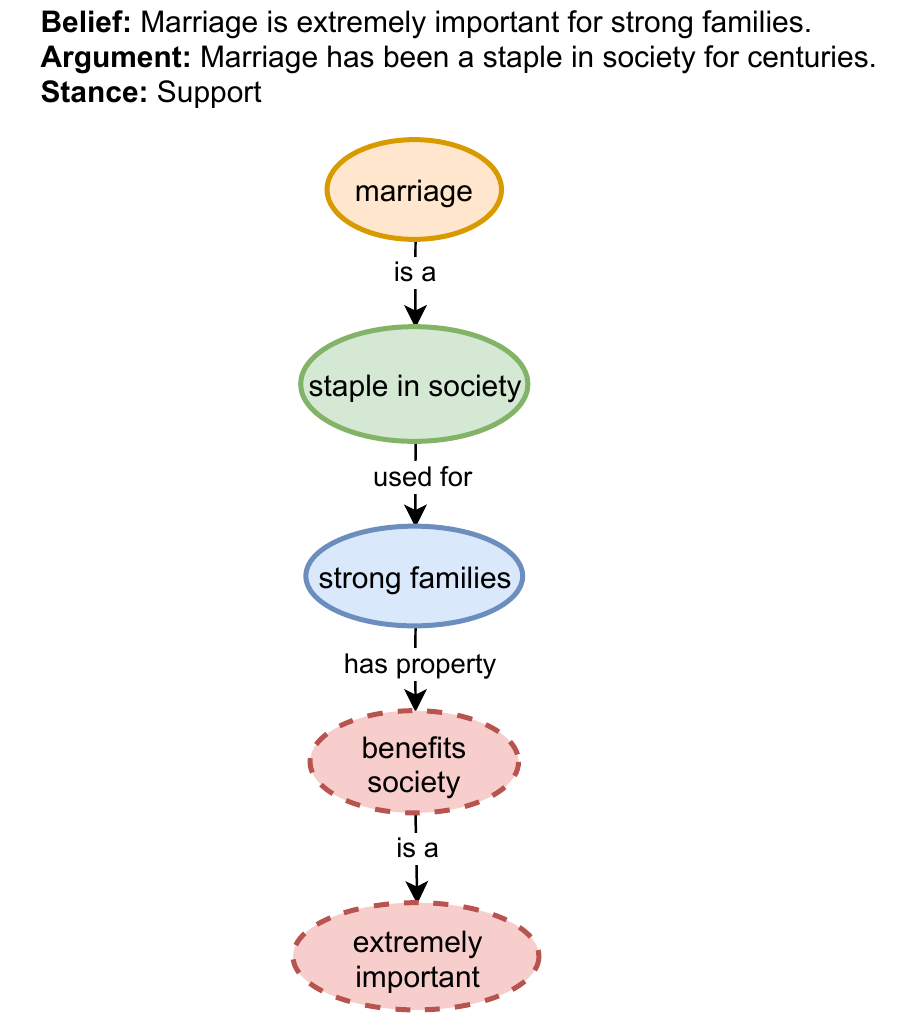}
     \caption{\label{fig:example8}Example 6} 
\end{figure}

\begin{figure}[tbh!]
	\centering
    \includegraphics[clip, width=\columnwidth]{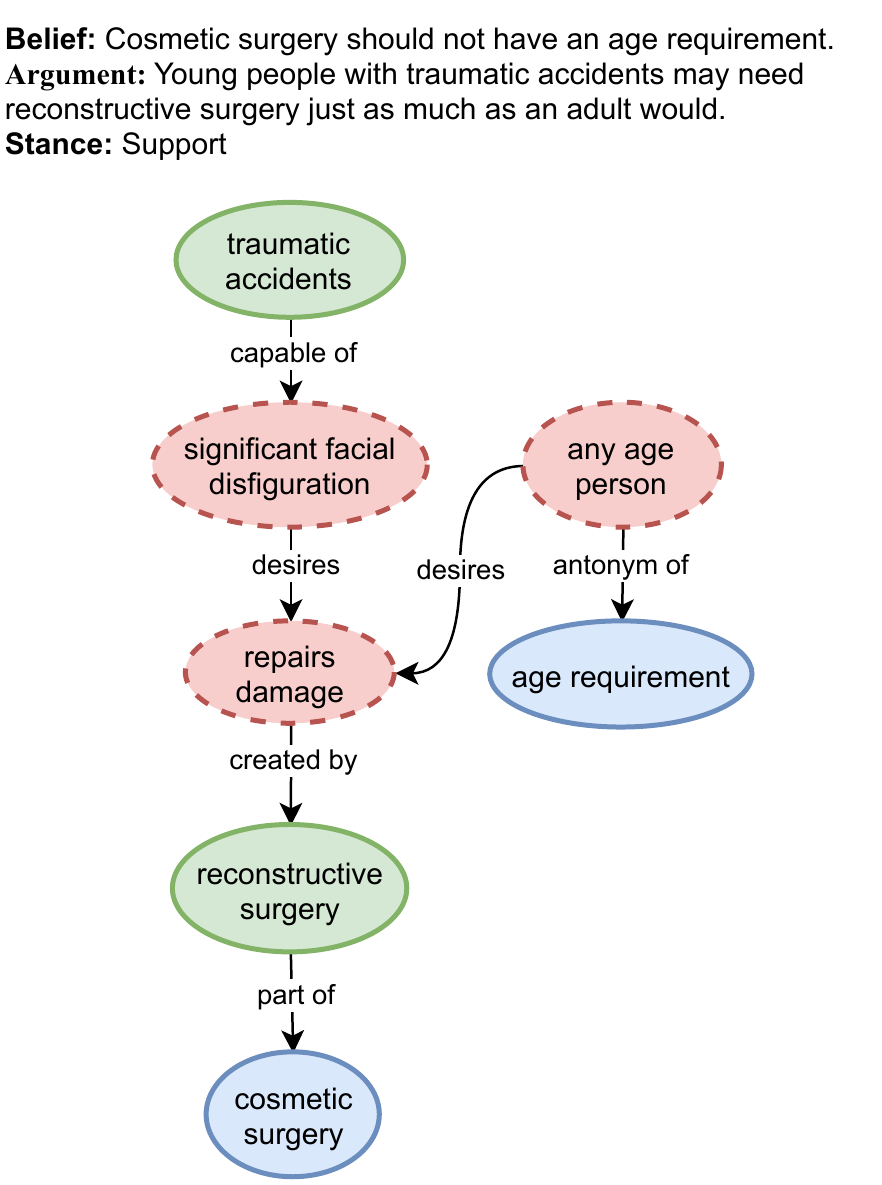}
     \caption{\label{fig:example9}Example 7} 
\end{figure}

\begin{figure}[tbh!]
	\centering
    \includegraphics[clip, width=0.85\columnwidth]{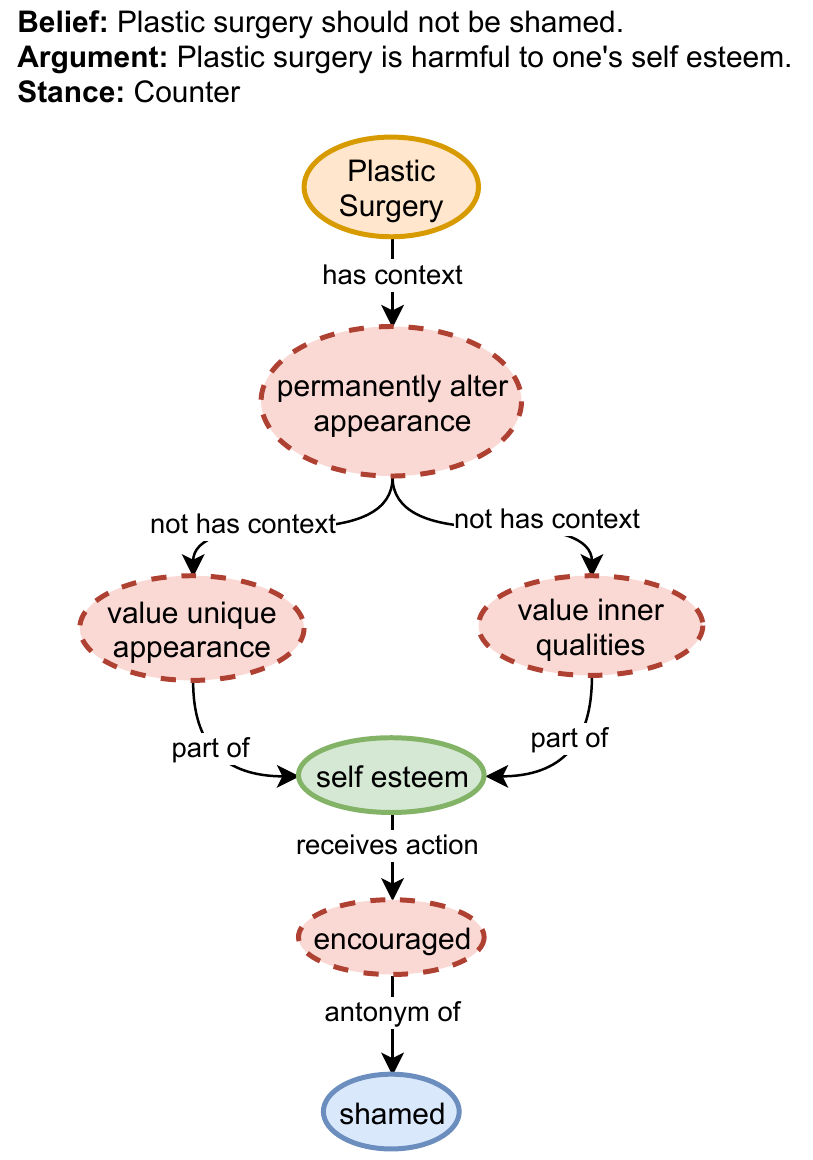}
     \caption{\label{fig:example10}Example 8} 
\end{figure}

\begin{figure}[tbh!]
	\centering
    \includegraphics[clip, width=\columnwidth]{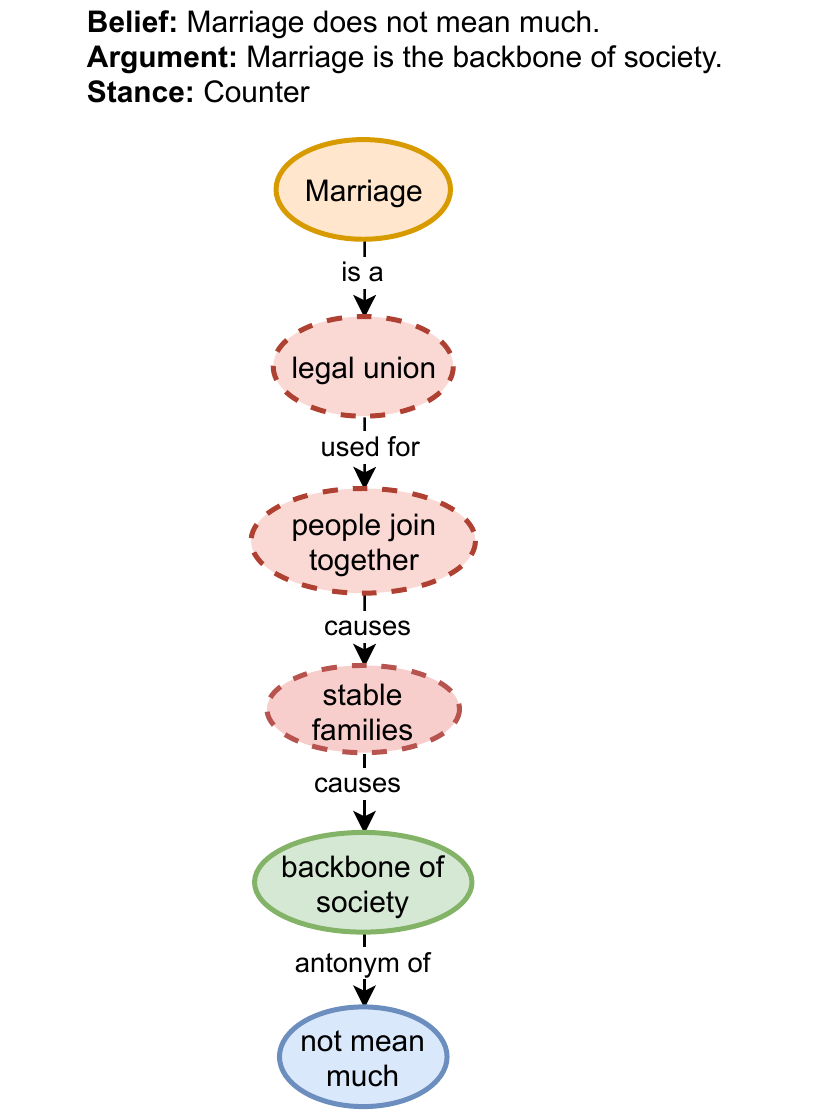}
     \caption{\label{fig:example11}Example 9} 
\end{figure}

We also show some randomly chosen examples from \data{} in Figures \ref{fig:example3}, \ref{fig:example4}, \ref{fig:example5}, \ref{fig:example6}, \ref{fig:example7}, \ref{fig:example8}, \ref{fig:example9}, \ref{fig:example10}, \ref{fig:example11}. Each example contains a belief, an argument, the stance and the corresponding commonsense explanation graph.

\end{document}